# LLM Agent Framework for Intelligent Change Analysis in Urban Environment using Remote Sensing Imagery

Zixuan Xiao[1], Jun Ma[1,*]

[1] Department of Urban Planning and Design, The University of Hong Kong, Hong Kong
[*] Corresponding author (Email address: junma@hku.hk)

**Abstract**

Existing change detection methods often lack the versatility to handle diverse real-world queries and the intelligence for comprehensive analysis. This paper presents a general agent framework, integrating Large Language Models (LLM) with vision foundation models to form ChangeGPT. A hierarchical structure is employed to mitigate hallucination. The agent was evaluated on a curated dataset of 140 questions categorized by real-world scenarios, encompassing various question types (e.g., Size, Class, Number) and complexities. The evaluation assessed the agent's tool selection ability (Precision/Recall) and overall query accuracy (Match). ChangeGPT, especially with a GPT-4-turbo backend, demonstrated superior performance, achieving a 90.71% Match rate. Its strength lies particularly in handling change-related queries requiring multi-step reasoning and robust tool selection. Practical effectiveness was further validated through a real-world urban change monitoring case study in Qianhai Bay, Shenzhen. By providing intelligence, adaptability, and multi-type change analysis, ChangeGPT offers a powerful solution for decision-making in remote sensing applications.

**Keywords:** Change Analysis; Large Language Model (LLM); Multi-modal Agent; Remote Sensing

## 1. Introduction

This section introduces the context and motivation behind the proposed research. It outlines the current challenges in change analysis using remote sensing imagery, identifies key research gaps, and summarizes the main contributions of this study.

## 1.1. Background and Motivation

Changes in remote sensing images provide valuable information, not only helping us identify current challenges, such as climate change impacts and urban expansion pressures, but also assisting in predicting future trends [1–4]. With advancements in remote sensing technologies, an increasing number of high-resolution images are now readily accessible, offering more detailed information about changes. This detailed information supports assessment and analysis of changes for future urban sustainable design, from land use planning to environmental monitoring [5].

Numerous case studies across different regions have demonstrated the utility of remote sensing in understanding and projecting spatiotemporal change. For example, Paul [1] used multi-temporal satellite data and GIS techniques to detect land use changes and predict urban growth in Habra I and II blocks, India. Göksel and Balçık [6] applied pansharpened SPOT-5 imagery to quantify land cover transitions in the Akdeniz District of Turkey, highlighting agricultural decline and urban expansion. Das and Dhorde [7] analyzed shoreline changes and mangrove retreat on India's Konkan coast, showing the interplay between coastal erosion and vegetation dynamics. In a broader environmental context, Boutallaka et al. [8] assessed current and projected land degradation sensitivity in Morocco's Ouergha watershed using a GIS–AHP framework under climate change scenarios. These studies collectively reinforce the importance of remote sensing for capturing and analyzing environmental and anthropogenic changes at various scales.

In addition, in land use planning, analysing historical change data enables planners to monitor development trends and prevent overdevelopment, ensuring balanced land use [9]. As for environmental monitoring, change assessments contribute to flood risk management [10] and the reduction of urban heat island effects [11,12]. Moreover, in the context of urban development, such analysis provide insights into the dynamics of urban sprawl, transportation, and infrastructure development, informing policy decisions for sustainable social and economic development. Overall, the information about changes derived from remote sensing images plays a crucial role in understanding and managing the sustainable interactions between urban environments and human societies [13–15].

To this end, there are many studies dedicated to developing precise change detection models [16,17]. These studies often focus on improving network design or specific adaptation for certain objects, such as buildings or roads, providing more accurate detection results, also referred to as change maps. However, these models typically consider only a single type of change, and the change maps they generate contain limited information, not sufficient for complex analyses required in practical application scenarios. For example, when analysing building changes, it is necessary to not only highlight the appearance or disappearance of buildings but also to calculate quantitative results and conduct comprehensive analysis on these results. A typical change detection model can only indicate whether a change has occurred, lacking detailed information about the changes of buildings, which cannot support further analysis. In addition, these models often lack the ability to respond intelligently with diverse demands of real-world problems. Consequently, these limitations hinder the accessibility of remote sensing interpretation, especially for various remote sensing application scenarios. Compared to conventional change detection models, which focus on generating accurate pixel-wise maps, our work shifts the focus toward multi-step, query-driven change analysis, addressing the gap between detection and actionable interpretation.

Recently, significant progress has been made in large language models (LLMs), making the development of intelligent systems increasingly feasible. These LLMs are trained with huge number of parameters by vast corpora, demonstrating remarkable ability in problem solving and code generating. However, due to lack of perception of images, these models suffer heavily from hallucination issues and struggle with precisely understanding images. Therefore, agents are designed to build with many specialized Vision Foundation Models (VFMs) to provide a fundamental connection between LLMs and images [18,19]. In the field of remote sensing, various agents [20,21] are introduced for adaption for special features of remote sensing images. Nevertheless, these remote sensing agents only consider basic demands rather than complicated change analysis. Since changes are critical elements in remote sensing, it is essential to harness the powerful capabilities of LLMs to develop a intelligent agent, specifically for change assessment and analysis in diverse application scenarios.

To address diverse challenges in real-world applications, this paper proposed a comprehensive agent framework that supports the development of intelligent agents for change assessment and analysis in an interactive manner. A hierarchical framework consists of two layers and a core planning navigator module is proposed to develop an interactive and intelligent agent, namely ChangeGPT, by integrating the capabilities of LLMs with a suite of specialized tools for analysing changes in remote sensing applications. This agent is designed to provide insights into complex changes in remote sensing images, significantly enhancing the capability and precision of change assessment and analysis in various practical scenarios.

As shown in Fig. 1, we compare the responses of three systems—traditional change detection models, a generic large language model (GPT-4), and our proposed ChangeGPT agent, when addressing an analytical query related to building changes. The traditional model generates a basic change map but fails to isolate specific object-level information (e.g., buildings), thereby requiring additional human interpretation and manual analysis. While GPT-4 attempts to address the query by generating Python code, it often produces incorrect answers due to hallucination issues. This limitation indicates extreme challenges when the model encounters queries beyond its existing domain knowledge, particularly those requiring specialized tools for complicated analysis.

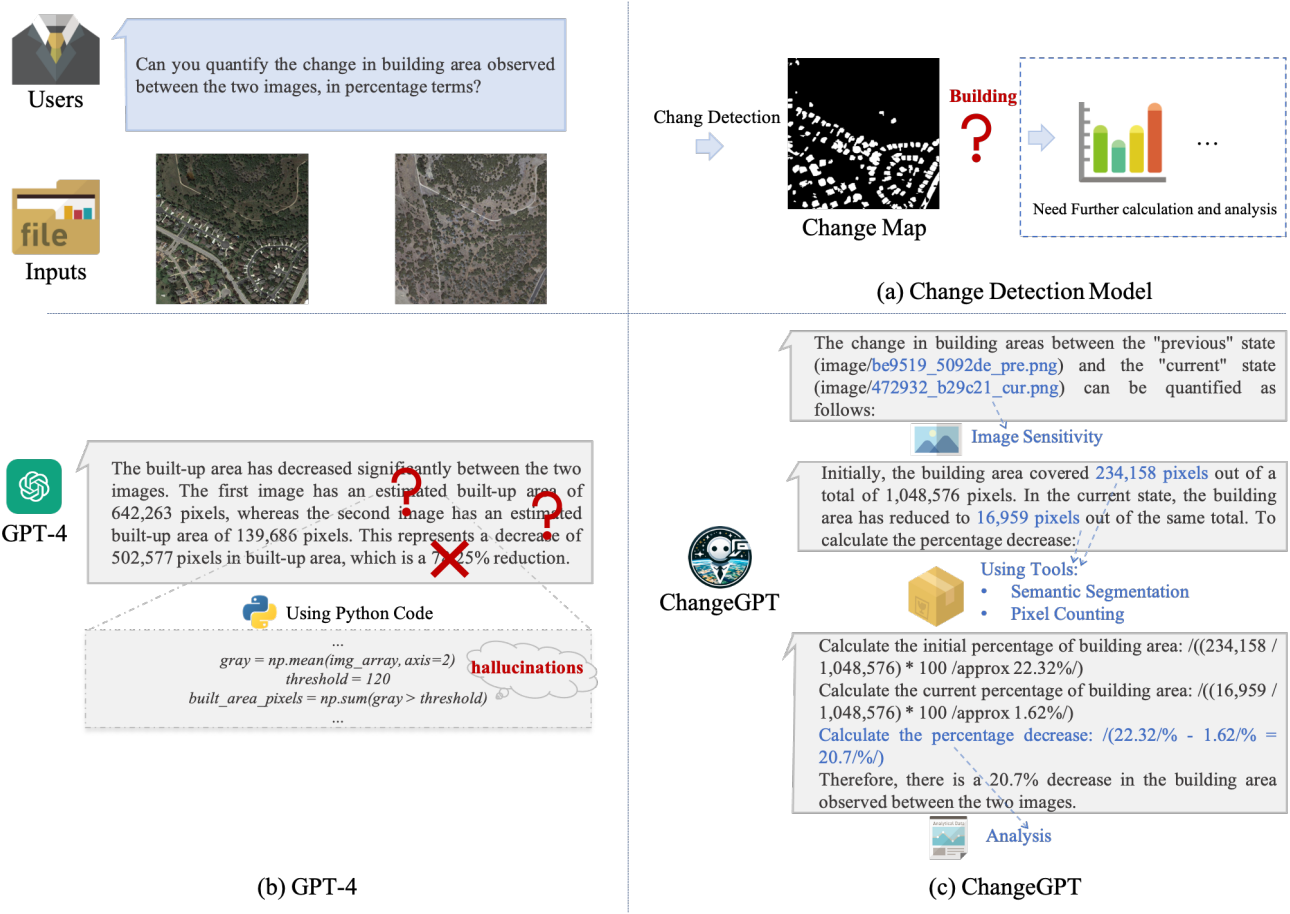

Fig. 1 Comparison of results among a traditional change detection model, a generic LLM (GPT-4), and the proposed ChangeGPT agent for answering building change analysis queries.

In contrast, ChangeGPT leverages its hierarchical agent architecture and planning navigator to parse the query, identify relevant imagery, and coordinate an appropriate sequence of tool-based operations. Following this, appropriate tools are selected to execute these intermediate steps. Ultimately, the agent provides an accurate analytical response. Overall, the agent enables a systematic and precise analysis of changes in remote sensing imagery, offering insightful and contextually relevant information tailored to specific changes.

### 1.2. Research Gaps and Contributions

Most current change detection approaches focus on pixel-level binary or semantic change maps, lacking the capacity to conduct further interpretation or support analytical decision-making. Similarly, while LLM-based remote sensing agents (e.g., RSGPT, SkyEyeGPT) [20,21] have begun addressing basic visual tasks, they lack explicit mechanisms for reasoning over temporal change, and typically operate on single-image inputs without structured planning or interpretability. Moreover, existing agent designs rarely support modular customization or transparent decision-tracking, which are essential for adapting to diverse application domains and building trust in urban scenarios. To address these limitations, this paper introduces ChangeGPT, an agent framework designed for query-driven remote sensing change analysis.

The contributions of the paper can be summarized as follows:

(1) Current change detection methods consider only a single type of change and lack the intelligence for further analysis. To address these limitations, we proposed a general agent framework for developing intelligent agents capable of querying diverse types of changes in real-world applications.

(2) Based on our framework, we developed ChangeGPT, an intelligent agent for analysing changes in remote sensing images. Additionally, we meticulously crafted a toolkit specifically for change analysis to alleviate the hallucination issues.

(3) A question dataset categorizing the questions by real-world application scenarios is developed. On this dataset, we conducted a comprehensive evaluation of our agent. Furthermore, we performed a case study to verify the practical effectiveness of our framework.

## 2. Related Work

This section reviews existing literature relevant to this work, including techniques in remote sensing change detection and the recent integration of large language models (LLMs) into autonomous agent systems.

### 2.1. Remote Sensing Change Detection

Traditional change detection methods for remote sensing images [22] can be grouped in three major categories: algebra-based, transformation-based, and classification-based techniques. Algebraic operations, such as image differencing, image regression, image ratioing, and change vector analysis (CVA), are performed on multi-temporal images for the change map in algebra-based techniques [23]. In transformation-based techniques, such as principal component analysis (PCA), multivariate alteration detection (MAD), gramm–schmidt orthogonalization (GS), and tasseled cap transformation (TCT), these renowned data reduction methods are employed to mitigate correlated information and accentuate variance within multi-temporal images [24].

Classification-based techniques include methods achieved by analysing multiple classification maps called post-classification comparison, or by utilizing a pre-trained classifier to directly assess the data named as direct classification [25]. These traditional models are susceptible to the influences of atmospheric conditions, changing seasons, satellite sensors, and solar elevations, influencing their accuracy of change detection.

With the development of deep learning, its exceptional capability for feature extraction has made profound enhancements in the domain of remote sensing change detection. Therefore, deep learning based remote sensing change detection methods significantly outperforms traditional approaches. Various architectures of deep learning, such as convolutional neural networks (CNNs) [26], recurrent neural networks (RNNs) [27], auto-encoders (AEs) [28], and Generative Adversarial Networks (GANs) [29], have been explored and developed. Recently, due to the advancement of computation capability, advanced large models like bi-temporal image Transformer (BIT), segment anything model (SAM), and contrastive language-image pre-training (CLIP) models are introduced in change detection as well to further improve the accuracy or reduce the reliance on extensive training data [30]. Although deep learning based change detection models can achieve high performance in detecting changes, they only consider single type of changes and lack of intelligence for further analysis on changes.

## 2.2. Large Language Models and Agents

These years, with the emergence of ChatGPT [31] and LLaMa 2 [32], LLMs, as intelligent models, have demonstrated amazing zero-shot and few-shot abilities compared with deep learning models [33,34] , such as language generation, emotion prediction, world knowledge, and reasoning, in almost every field in daily life. These models are trained on extensive corpora and consist of a vast number of parameters, which, while enhancing their performance, also lead to high computational costs for fine-tuning. In some ways, prompt engineering has emerged as a popular method to leverage the full potential of LLMs efficiently [35]. Through designing specific prompts, this method aims to guide the responses of the models in a desired direction without the need for additional fine-tuning [36]. Consequently, the careful design of prompt formats, organizing them as preliminaries and principles in a system, is crucial for ensuring the effective and reliable use of LLMs in practical applications.

However, no matter how to design the prompts, LLMs are not able to directly understand other format of information, such as images and videos. To this end, agents are designed with multiple vision foundation models (VFMs) to empower LLMs the ability to accurately respond with queries concerning images understanding. In agent development, there are some studies designing agents to understand images with models trained with general images [37–39]. These models often face significant challenges caused by the special features in remote sensing images, resulting in bad performance.

In response, more recent works have introduced agents tailored to remote sensing applications [40–42]. These systems often support foundational tasks such as scene understanding, land cover classification, or simple image captioning. However, they are primarily designed for single-image reasoning and lack the capacity to reason across bi-temporal inputs, a core requirement for change detection and interpretation. Another key limitation of these agents lies in their lack of explainability. Most existing remote sensing LLM agents produce outputs directly from prompt-to-response generation without exposing intermediate reasoning steps, tool usage paths, or execution logic. This opaqueness makes it difficult for users to understand, verify, or trust the rationale behind analytical conclusions, particularly in urban analysis scenarios. By contrast, ChangeGPT is explicitly designed with explainability in mind, maintaining structured reasoning records, including query decomposition steps, selected tools, intermediate outputs, and execution order.

A notable attempt to build a change-focused agent is Change-Agent [43], which trains a multi-modal LLM using a specialized dataset (LEVIR-MCI) composed of change masks and captions for buildings and roads. While effective within its domain, this approach is inherently limited by its supervised training setup, both in task diversity and generalization capability. Its reliance on end-to-end learning restricts flexibility, as adapting to new object classes, change types, or analysis styles would require retraining or substantial dataset expansion. In contrast, our proposed ChangeGPT framework takes a modular and reasoning-driven approach. Rather than relying on joint multi-modal training, ChangeGPT employs a hierarchical agent structure that coordinates a set of independently developed and interchangeable VFMs through an LLM-based planning module. This design empowers the agent to handle a wide variety of analytical change queries, including quantitative, comparative, and localized tasks, without the need for model retraining. It also provides enhanced adaptability, allowing users to flexibly select or update tool models according to their own data or application domains.

## 3. Method

This section details the proposed methodology, including the problem definition and the architecture of the ChangeGPT agent framework. It elaborates on the design of the planning navigator and the toolkit components that support intelligent decision-making in remote sensing change analysis.

### 3.1. Problem and Model Definition

Remote sensing change detection is primarily concerned with identifying changes within remote sensing imagery captured across different time intervals. Specifically, this research focuses on RGB remote sensing images and comparisons between these two images: one taken before the change (referred to as the "previous image") and one taken after the change (referred to as the "current image"). In our framework, the planning navigator acting as a pivot is denoted as $\mathcal{N}$. The backend Large Language Model (LLM) is denoted as $\mathcal{L}$ and can be either GPT or other LLMs.

Assuming that in round $h$, the agent responds with an answer $\mathcal{A}_h^i$ based on vision foundation model (VFM) $i$:

$$\mathcal{A}_h^i = \mathcal{L}(\mathcal{N}(P), \mathcal{N}(F), \mathcal{N}(H_{<h}), \mathcal{N}(Q_h), \mathcal{N}(\mathcal{R}_h^{<i-1}), \mathcal{N}\left(F(\mathcal{A}_h^{i-1})\right)) \qquad (1)$$

$P$ represents both preliminaries and principles as prompts that define role and constraints for LLM. These guidelines are essential for tailoring the responses, ensuring that its output aligns with the intended framework and objectives. The formulation of $P$ can be expanded and broken down into several key components, each serving a specific purpose within the system. The mathematical representation of $P$ is as follows:

$$P = (P_{prefix}, P_{image}, P_{reference}, P_{format}, P_{suffix}) \qquad (2)$$

Each part in this decomposition plays a critical role in shaping the agent's decision-making process and ensuring its output remains aligned with the task objectives.

The first part, $P_{prefix}$, represents the Role Definition Prompts. These prompts provide a foundational introduction to the agent's role within the framework, outlining its tasks and defining the constraints it must adhere to when responding. Next, $P_{image}$, which stands for Unique Image Identification, is critical in helping the agent correctly identify and differentiate between the images being processed. The third component, $P_{reference}$, refers to Reference Descriptions, which provide important metadata related to the images being analyzed. This includes image paths, storage locations, or other forms of referencing that enable the agent to access the images accurately. In addition to the reference information, $P_{format}$, which stands for Format Instruction Prompts, is used to ensure that the agent's output adheres to the required format. Consistent formatting is crucial for integrating the agent's responses with other parts of the framework or for providing outputs that can be easily interpreted and used by downstream applications. Finally, the $P_{suffix}$, which represents Suffix Prompts, plays a vital role in reinforcing the agent's operational constraints and highlighting any critical considerations that must be taken into account during response generation.

$F$ signifies one of models from the remote sensing VFMs selected by the agent, indicating the integration of remote sensing visual analysis capabilities into the process. $H_{<h}$ is dialogue history before the current round. The query from users at $h-th$ round is denoted as $Q_h$ and the $\mathcal{R}_h^{<i-1}$ is the reasoning and planning historic results from previous $i-1$ VFMs at $h-th$ round. Finally, the intermediate answer is denoted as $\mathcal{A}_h^{i-1}$.

$$H_{<h} = \{(Q_1, \mathcal{A}_1), (Q_2, \mathcal{A}_2), \ldots, (Q_{h-1}, \mathcal{A}_{h-1})\} \quad (3)$$

### 3.2. Agent Framework

The general framework is depicted in Fig. 2. Inspired by telecommunication network protocol designs, our proposed framework is hierarchical and consists of three parts: an application layer, a core planning navigator module and an executing layer. This hierarchical framework provides a standard and modular design guide for building our agent on change assessment and analysis. The layer division follows the principle that different layers with different functionalities focus on specific tasks. As for data flow, similar to telecommunication network protocol, from up to down, the data, both images and queries, are processed step by step, following a standard principle as well.

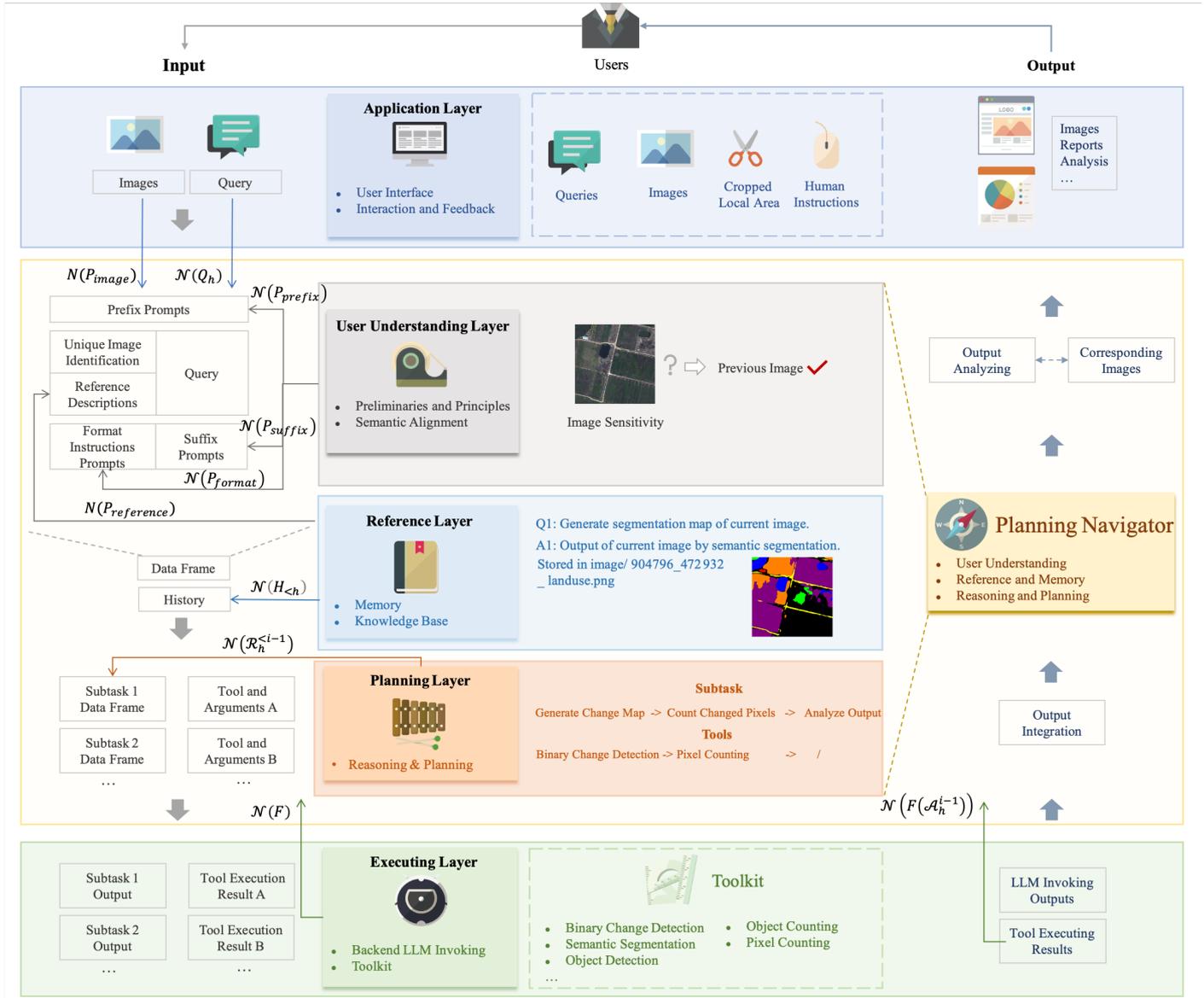

Fig. 2 General framework of agent for change assessment and analysis in remote sensing images. This framework follows a hierarchical design with five layers and a core module.

Specifically, the application layer serves as a user interface (UI). Apart from common query and answer demonstration, it enables human instructions, such as image cropping. It is important for an agent to understand where the area that users refer to is. Only language queries are not enough to explain the correct message, especially for analysing changes in large VHR remote sensing images. Therefore, human instructions are necessary. On the other hand, despite the strong ability of LLM, a comprehensive analysis or even report cannot be likely generated in a single round. In this layer, the UI support multi-round queries, users can propose feedbacks for answers and acquire the outputs in the same place as well.

In the middle, the core module, namely planning navigator, is formed by three layers (user understanding layer, reference layer, and planning layer) in our framework. The planning navigator is pivotal to guide the workflow to coordinate between the LLM and the toolkit to address the designated tasks effectively. It is endowed with three specific functions: user understanding, reference and memory, and reasoning and planning.

Particularly, in the user understanding layer, some preliminaries and principles are pre-defined and function as a former of prompts. Moreover, the specific image sensitivity ability is designed to make sure the agent is well semantically aligned with users. For example, the simplest and most important part is to ensure that the agent is able to differentiate between previous and current images. In the reference layer, some reference data is introduced for quick responses or precisely locating of some intermediate outputs, such as files and images. The history, storing details of previous queries, reasoning steps, responses, and actions, can also be retrieved as a reference for future decisions. In the planning layer, facing complex queries, a systematic process of reasoning and planning is necessary, structuring well-organized steps for the final solution.

The Reference Layer in Planning Navigator is key to the explainability of ChangeGPT. The Reference Layer stores not only past user queries, but also intermediate reasoning outputs, tool invocation details, and subtask execution logs. These elements are accessible during subsequent reasoning steps, enabling the agent to build upon and contextualize new decisions using historical knowledge.

More importantly, the Planning Navigator generates an explicit chain-of-thought execution plan for each complex query. These plans are made transparent to the user, either through structured summaries or directly returned execution steps, allowing human operators to understand the logic, verify correctness, or intervene if necessary.

In the executing layer, the subtasks planned by the backend LLM are organized into a pool, each associated with specific tools from the toolkit suited to their completion. The toolkit contains multiple remote sensing

VFMs, particularly with detailed descriptions of when and how to invoke them so that the agent is able to select the right tool and needed arguments.

To sum up, this general framework ensures the agent to act in a coherent and efficient way, allowing for the dynamic adaptation and handling of complex queries asking for analysis or reports in changes of remote sensing images.

**3.3. Planning Navigator**

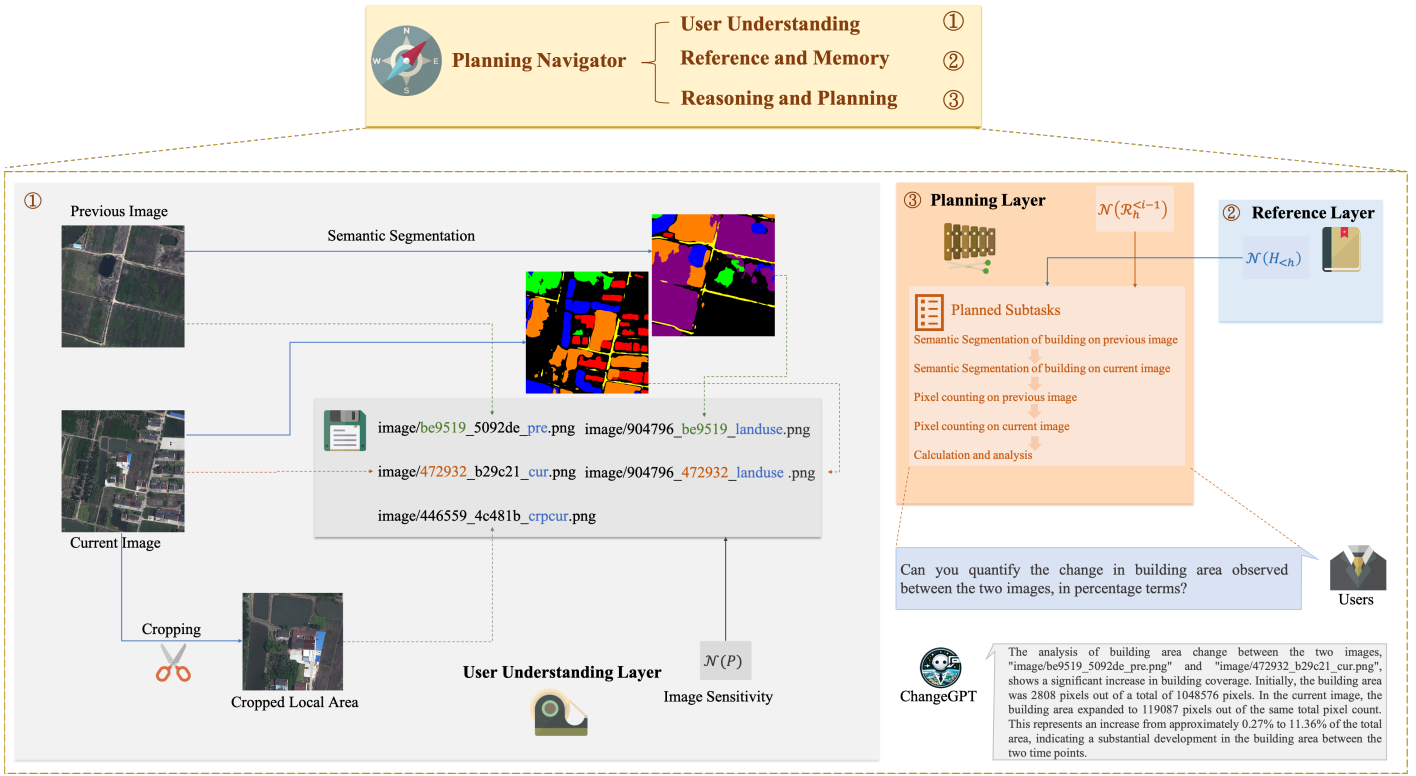

Fig. 3 Three functions of planning navigator: user understanding, reference and memory, and reasoning and planning.

As mentioned in the Section 3.2, we will introduce the planning navigator in our framework in detail. The functions of the planning navigator are illustrated in Fig. 3, user understanding, reference and memory, and reasoning and planning.

In the complex framework, in order to effectively constrain backend LLM to concentrate on user queries and efficiently manage various VLMs, preliminaries and principles need to be systematically defined and transferred as a form of prompts understandable by the LLM. Specifically, the role of the agent is defined to emphasize that it is designed to solve queries concerning changes in remote sensing images. Apart from role of the agent, answer format and constrains of the agent are clarified as well, aiming to designed to guide the its operation and alleviate hallucinations to the maximum extent.

Beyond principles, image sensitivity is also a specifical ability of the agent. With this ability, as depicted in Fig. 3, when receiving an image as input, the agent stores it locally under a unique identifier that reflects its temporal context. For example, previous image is labelled with a suffix 'pre', as in 'be9519_5092de_pre.png', where 'pre' denotes the status that it is captured before change. Similarly, the cropped area from a previous image would include 'crppre' in its naming. Additionally, processed outcomes, such as those from remote sensing semantic segmentation, are assigned names like '904796_be9519_landuse.png'. Here, '904796' serves as a unique identifier and 'be9519' is denoting the parent of this image to keep a track. And 'landuse' indicates the type of processing applied, in this case, being semantic segmentation. This meticulous naming scheme not only ensures the traceability of each image and the cropped one and even its derivatives but also enhances the system's ability to manage and utilize these images precisely for semantic alignment with users. This enables the agent to respond with the right image that user refers to. Furthermore, interactions and accessibility with VFMs and the reasoning format are governed by specific rules to mitigate the risk of inaccuracies from hallucinations, thereby maintaining reliability of the agent.

Besides the function for user understanding, in the phase of completing, the memory storage and retrieval are important abilities as well to refer to historic information and alleviate hallucinations. In addition, reasoning and planning also plays a crucial part in maintain the agent. Facing a complex analytical query, the agent should be capable of reasoning the internal logic of the query and planning to divide the whole tasks into small subtasks. For example, when asked "Can you quantify the change in building area observed between the two images, in percentage terms?", the agent will reason with using semantic segmentation VFM to acquire the size and do comparison, and then divide into segmenting firstly, counting pixels next, and finally calculating. To sum up, the whole module, combining careful design with robust operational guidelines, navigates the agent in analysing the complex changes in remote sensing images accurately and efficiently.

## 3.4. Toolkit

The toolkit in our framework contains necessary VFMs to solve different diverse queries of changes in remote sensing images. A summary of the backend remote sensing VFMs is illustrated in Table 1.

Table 1. Tools in toolkit.

| Tool Type | Tool Name | Method | Dataset |
|---|---|---|---|
| Deep Learning | Binary Change Detection | SAM & CLIP | / |
| | Image Captioning | BLIP | BLIP |
| | Scene Classification | ResNet | AID |
| | Semantic Segmentation | DCSwin | LoveDA |
| | Object Detection | YOLOv5 | DOTA |
| | Object Counting | YOLOv5 | DOTA |
| Basic Calculation | Pixel Counting | / | / |
| | Whether Change | / | / |

There are two types of models chosen in the toolkit. The models in the first type are trained by deep learning, including binary change detection, image captioning [44], scene classification [45,46], semantic segmentation [47,48], object detection [45,33,34,49], and object counting. These models are specifically trained with remote sensing datasets or adapted for remote sensing domain, enabling the agent to interpret complex image data and analyse changes over time. On the other hand, there are two basic calculation models, pixel counting, and whether change. They grant our agent with precise calculation ability for certain analysis.

The analytical prowess of the agent in addressing varied queries related to changes in remote sensing images is directly supported by this toolkit, including both deep learning models and basic calculation models. This comprehensive suite of analytical tools provides the agent with the necessary capabilities to process a wide range of queries with precision and efficiency.

The models listed in the toolkit (Table 1) are chosen based on their reproducibility, proven performance on remote sensing tasks, and ease of integration for demonstration purposes. However, they are not fixed components. The toolkit is designed to be modular and extensible. Alternative models, such as DETR [50] or Faster R-CNN [51] for object detection, can be readily integrated depending on the specific requirements of a deployment scenario, such as object scale variation or inference speed constraints.

## 4. Experiment

This section presents the experimental setup and evaluation results of the proposed framework. It covers the datasets used, quantitative performance metrics, comparative analysis, and case studies that demonstrate the capabilities of the proposed approach.

### 4.1. Remote Sensing Data Source, Question Dataset and Evaluation Metric

Since the proposed framework focuses on responding with diverse user queries about changes accurately, we have developed a strategic question dataset according to real-world scenarios. As shown in Table 2, there are mainly four types of questions in this paper. The first one is 'Whether' question, indicating user queries about whether there exists change or not. The second type is 'Size' question to quantify the changes in pixel-level. Whereas, 'Number' question is to quantify the changes in object-level as the third type. The last type is for qualitative analysis about class change denoted as 'Class' type, such as the changes from farmland to building in an image.

Table 2 Question dataset in experiment.

| Question Type | Subtype | Number | Example |
|---|---|---|---|
| Whether | / | 15 | Is there a discernible difference between the images indicating changes? |
| Size | Basic | 10 | Estimate the percentage of the changed area relative to the total image size. |
| | Certain Class | 15 | What proportion of the water bodies has increased or decreased in size, expressd as a percentage? |
| | Local Area | 10 | In the localized area I have cropped, what percentage of the area has undergone changes? |
| | Analysis | 10 | Compare the pixel changes in Buildings and Roads (indicative of urban development) to Farmland, and quantify the shift to assess urban sprawl. |
| Number | Basic | 15 | Between the previous and current images, has there been an increase or decrease in the number of ships? |
| | Local Area | 10 | For the cropped area, can you calculate the change in the number of planes between the two images? |
| | Comparison | 20 | Compare the change in the number of storage tanks to the change in the number of harbors between the previous and current images and determine which category experienced a greater change in number? |
| Class | Whole Image | 20 | What class covered the entire area of the previous image, and to what class does the entire area belong now? |
| | Local Area | 15 | In the area I have cropped from the whole image, what was the class before the change occurred? |

In some question types, such as 'Size' type, there are also some subtypes of questions. As illustrated in Table 2, in 'Size' type, 'Basic' subtype denotes elementary questions about changed size in quantities not constrained for certain class and concentrating on the whole image. 'Certain Class' and 'Local Area' are queries about changed size for certain class and for local area respectively. 'Analysis' subtype involves more complex queries that compare or generate analytical reports, such as some analysis about urbanization plans. Similarly, for the 'Number' category, 'Basic' and 'Local Area' stand for queries about quantifying the changed number for certain object in the whole images or some local area of the images respectively. 'Comparison' questions focus on comparing changed number of different objects. The two subtypes in 'Class' category,

'Whole Image' and 'Local Area', targeting at asking the changed information about class based on whole images or some local area.

To support the proposed question-driven evaluation framework, we utilized remote sensing images as samples from public remote sensing datasets, LEVIR-CD datasets [52], widely used in remote sensing change detection research.

The LEVIR-CD dataset is composed of large-scale building change pairs with a spatial resolution of 0.5 m/pixel, consisting of RGB images collected from diverse urban environments. We used LEVIR-CD image samples as test cases for evaluating quantitative queries in our question dataset.

For each question in the dataset, a reference answer was manually created and verified by the research team. For analytical or qualitative queries, such as those involving class transitions or comparative size analysis, the responses were generated collaboratively by team members with expertise in urban studies and remote sensing interpretation to ensure semantic accuracy and consistency. This process ensures that the reference answers serve as a robust basis for evaluating the agent's reasoning performance through the Match metric.

Besides the question dataset, we designed the evaluation metrics inspired by evaluation systems of models in machine learning. The evaluation system is structured to assess the performance of our agent in two dimensions. On the one hand, we aim to evaluate the ability of the agent in selecting the appropriate tools from the toolkit. It is vital because the correct tool selection directly impacts its ability to understand the queries and complete tasks effectively, thus influencing the overall solution it generates. On the other hand, the performance in providing correct answers to the queries will be evaluated, focusing on generating the accurate and corresponding outcomes with the queries.

Similar to the evaluation systems in machine learning, we define the Precision $\mathcal{P}$ and Recall $\mathcal{R}$ in our evaluation system:

$$\mathcal{P} = \frac{\mathcal{N}(RT)}{\mathcal{N}(TT)} \qquad (4)$$

$$\mathcal{R} = \frac{\mathcal{N}(RT)}{\mathcal{N}(NT)} \tag{5}$$

Both Precision $\mathcal{P}$ and Recall $\mathcal{R}$ serve as key metrics for evaluating the tool selection efficacy of the agent. Each one of them provides insight into different aspects of tool selection performance. Precision $\mathcal{P}$ can be calculated by the number of right tools selected in answering this query $\mathcal{N}(RT)$ divided by the total number of tools the agent used $\mathcal{N}(TT)$. This metric indicates the efficiency in selecting tools, highlighting the ability of choosing more relevant tools over irrelevant ones. Whereas, Recall $\mathcal{R}$ indicates that the agent selects most of the correctly needed tools, calculated by $\mathcal{N}(RT)$ divided by the total number of tools actually needed in answering this query. This metric reflects the ability to identify all necessary tools for query solution, ensuring that not any needed tools are omitted.

In addition, in order to evaluate the general question answering performance of the agent, we proposed Match $\mathcal{M}$, which is defined as follows:

$$\mathcal{M} = \frac{\mathcal{N}(RA)}{\mathcal{N}(TQ)} \tag{6}$$

We utilize the number of right answers $\mathcal{N}(RA)$ divided by the number of total queries $\mathcal{N}(TQ)$ to calculate Match $\mathcal{M}$. Particularly, different from the previous two metrics, this metric focuses on the results of the agent by assessing the matching rate of all generated solutions in our question dataset. It only considers whether the output solution can effectively address the query, irrespective of whether it contains irrelevant tools. Therefore, both metrics on evaluating tool selection and question answering performance are necessary.

Additionally, the traditional segmentation metrics such as IoU or F1-score are not directly applicable in our evaluation setting, as our framework does not aim to optimize pixel-level predictions, but rather focuses on evaluating end-to-end analytical accuracy in response to task-specific queries.

### 4.2. Quantitative Performance and Discussion

In this section, based on our built question dataset, we will demonstrate the quantitative performance of our agent with previously introduced evaluation metrics. On the one hand, we test our agent with three different

LLM backends, GPT-3.5-turbo, Gemini Pro 1.0 [53], and GPT-4-turbo respectively and show the quantitative performance in Fig. 4. GPT-4-turbo is currently recognized as the most sophisticated LLM, with exceptional reasoning capabilities. For comparison, we also considered GPT-3.5-turbo, another model in the GPT series. Additionally, the experiment included Gemini Pro 1.0 as well, a LLM developed by Google to explore the performance against a different network design. In Fig. 4 (a) and Fig. 4 (b), the performance of tool selection is presented using overall precision and recall metrics.

Additionally, the detailed precision and recall for different subtypes within each question type are shown in the same figure. The figures reveal that the overall precision and recall performance are similar among the LLMs for 'Whether', 'Number', and 'Class' question types and their subtypes, indicating that all models have a nearly equal ability in tool selection for these questions. However, when addressing 'Size' type questions, GPT-4-turbo, as the backend in our agent, demonstrates superior capability, particularly in the 'Analysis' subtype. This suggests that the other two backends struggle to select accurate and necessary tools when dealing with complex and analytical queries.

Fig. 4 (c) presented the overall and detailed Match value for the tested queries. GPT-3.5-turbo performs well with 'Whether' and 'Class' type questions. However, it struggles with complex queries, such as the 'Comparison' subtype within 'Number' type questions. Both GPT-3.5-turbo and Gemini Pro 1.0 fail to respond effectively to 'Size' questions, particularly in the 'Analysis' subtype since they are unable to select the necessary tools, as indicated by their poor precision and recall performance. While they perform relatively well in tool selection for the other subtypes of 'Size' questions, their overall question-answering performance is hindered by limited ability to synthesize and analyze intermediate results. In conclusion, in overall question-answering performance, the agent with the GPT-4-turbo backend outperforms those with Gemini Pro 1.0 and GPT-3.5-turbo backends.

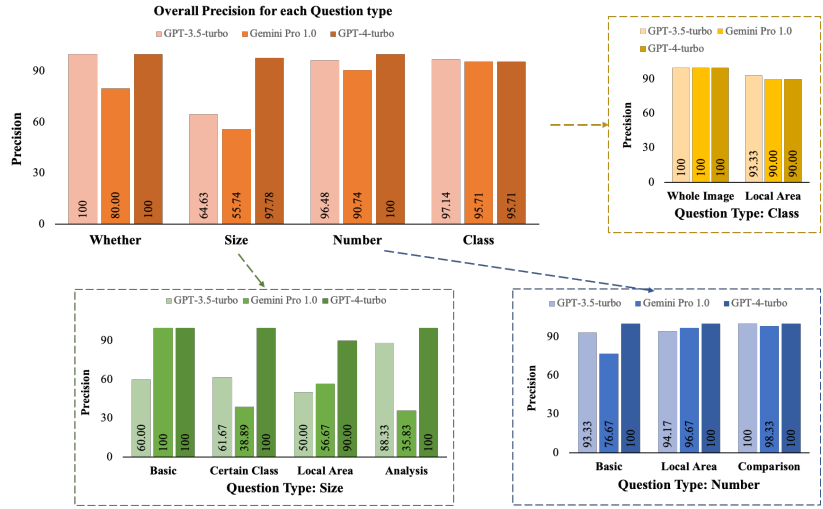

(a) Overall and Detailed Precision

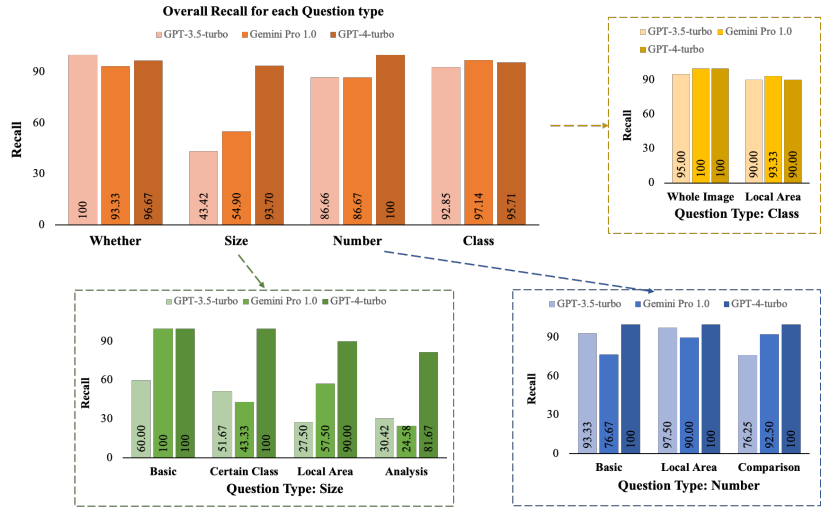

(b) Overall and Detailed Recall

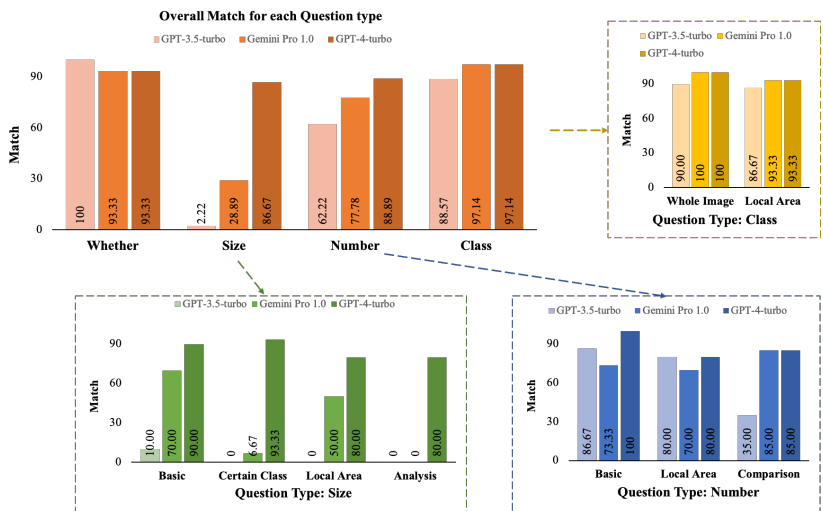

(c) Overall and Detailed Match

Fig. 4 Quantitative performance of ChangeGPT agent on tool selection and question answering.

Additionally, beyond evaluating performance based on question type, we categorize each question in our dataset into three levels of difficulty: Easy, Medium, and Difficult, according to the number of tools required to answer the question. 'Easy' questions require only one tool for the solution while 'Medium' questions require two tools. 'Difficult' questions require more than two tools. The results are presented in Table 3. From the table, it is evident that as the difficulty of the questions increases, performance deteriorates in both tool selection and question answering. The agent with the GPT-4-turbo backend consistently outperforms those with Gemini Pro 1.0 and GPT-3.5-turbo backends in question answering. Although the GPT-3.5-turbo backend performs better in tool selection than the Gemini Pro 1.0 backend, the latter achieves superior outcomes in final question answering. In conclusion, the GPT-4-turbo backend agent demonstrates the highest overall performance, especially when the complexity of the questions increases, indicating its robust capability in handling more challenging tasks.

Table 3. Performance according to difficulty of questions.

| LLM | Questions | Tool Selection | | Question Answering | | |
|---|---|---|---|---|---|---|
| | | Precision | Recall | Match | Δ vs Baseline | p-value |
| GPT-3.5-turbo | Easy | 93.75 | 93.75 | 93.75 | / | / |
| | Medium | 86.04 | 84.46 | 71.62 | / | / |
| | Difficult | 85.67 | 57.08 | 14.00 | / | / |
| | **Total** | **86.78** | **75.73** | **53.57** | **+0.071** | **0.253** |
| Gemini Pro 1.0 | Easy | 96.88 | 100 | 100 | / | / |
| | Medium | 88.29 | 91.22 | 66.22 | / | / |
| | Difficult | 61.16 | 56.42 | 36.00 | / | / |
| | **Total** | **79.58** | **79.79** | **68.57** | **+0.214** | **<0.001** |
| GPT-4-turbo | Easy | 100 | 100 | 100 | / | / |
| | Medium | 97.97 | 97.30 | 91.89 | / | / |
| | Difficult | 100 | 96.33 | 86.00 | / | / |
| | **Total** | **98.21** | **96.54** | **90.71** | **+0.443** | **<0.001** |

To rigorously evaluate the performance improvement of each LLM backend relative to the baseline (vanilla GPT-4-turbo), we conducted statistical significance tests using McNemar's test for the Match metric. Note that the baseline model is evaluated solely on its query-analyzing capability (Match) and does not involve tool selection. Consequently, it does not have Precision and Recall metrics, which specifically measure the agent's performance in tool selection. Table 3 presents the absolute differences ($\Delta$) compared to the baseline, and their corresponding p-values. The GPT-4-turbo backend agent demonstrates a statistically significant improvement with a large margin ($\Delta = +0.443$, $p < 0.001$). Similarly, the Gemini Pro 1.0 backend agent also exhibits a significant performance increase over the baseline ($\Delta = +0.214$, $p < 0.001$). Although the GPT-3.5-turbo backend agent shows a slight numerical improvement ($\Delta = +0.071$), it is not statistically significant ($p = 0.253$). To further evaluate the performance distinctions among different LLM backends, we conducted statistical significance testing based on the collected Precision, Recall, and Match scores across all 140 queries for each backend. One-way ANOVA analysis (Table 4) indicates that the observed differences among GPT-3.5-turbo, Gemini Pro 1.0, and GPT-4-turbo are statistically significant ($p < 0.001$) across all three metrics. Moreover, Tukey's HSD test (Table 5) confirms that GPT-4-turbo significantly outperforms both Gemini Pro 1.0 and GPT-3.5-turbo across all three evaluation metrics (Precision, Recall, and Match). While GPT-3.5-turbo also shows significantly higher Match performance than Gemini Pro 1.0, the differences in Precision and Recall between them are not statistically significant.

In addition, we conducted a detailed error analysis to explore the underlying causes of incorrect answers (i.e., queries with zero match score). The errors are categorized into four types, based on a detailed analysis of tool invocation behaviors and their alignment with query intent:

- Misunderstood Query: This occurs when both precision and recall are equal to 1, meaning the system selected all the necessary tools and no incorrect ones. Despite this, the final answer is wrong, indicating that the agent fundamentally misunderstood the semantics or intent of the question, even though the execution plan was technically correct.

- Insufficient Tools Used: When recall is low but precision is high, the tools that were selected are mostly correct, but not all necessary tools were included. This suggests that the agent partially understood the query but failed to plan a complete pipeline, missing required analytical steps.
- Incorrect Tools Used: When precision is lower than recall, the agent included irrelevant or incorrect tools, even though it may have captured many necessary ones. This typically reflects poor judgment in tool selection, likely due to a misinterpretation of image content, query constraints, or required reasoning operations.
- Too Complex: When both precision and recall are zero, it indicates the system failed to identify or apply any tools, suggesting that the query exceeded the system's capability to decompose or plan a valid execution path. This reflects the current limitation of the agent in dealing with highly abstract or compound queries.

The distribution of these error types is shown in Fig 5. From the figure, we observe that when our agent consider GPT-4-turbo as backend, it commits far fewer errors overall and most of them arise from subtle misinterpretations, suggesting its reasoning capability is well-developed. In contrast, GPT-3.5-turbo backend exhibits a large number of tool usage errors, mostly due to insufficient or incorrect tool selection. Gemini Pro 1.0 as backend, on the other hand, struggles with both tool selection and reasoning, with a considerable portion of its errors attributed to complexity. These analyses collectively affirm that when GPT-4-turbo is selected as backend, it not only achieves the best quantitative results but also exhibits more interpretable and reliable behavior when handling complex multi-step change queries.

Table 4 ANOVA Test Results for Metric Differences among LLM Backends.

| Evaluation Metircs | F-value | p-value | Conclusion |
|---|---|---|---|
| Precision | 16.99 | 8.08e-08 | Significant |
| Recall | 16.62 | 1.13e-07 | Significant |
| Match | 26.56 | 1.39e-11 | Significant |

Table 5 Tukey HSD Pairwise Comparison of LLM Performance.

| Comparison | Metric | Mean Diff | p-value | Significant or not |
|---|---|---|---|---|
| GPT-4-turbo vs GPT-3.5-turbo | Precision | +0.1173 | 0.0007 | √ |
| GPT-4-turbo vs Gemini Pro 1.0 | Precision | +0.1815 | <0.0001 | √ |
| GPT-3.5-turbo vs Gemini Pro 1.0 | Precision | -0.0643 | 0.1051 | × |
| GPT-4-turbo vs GPT-3.5-turbo | Recall | +0.1967 | <0.0001 | √ |
| GPT-4-turbo vs Gemini Pro 1.0 | Recall | +0.1545 | <0.0001 | √ |
| GPT-3.5-turbo vs Gemini Pro 1.0 | Recall | +0.0423 | 0.4679 | × |
| GPT-4-turbo vs GPT-3.5-turbo | Match | +0.3714 | <0.0001 | √ |
| GPT-4-turbo vs Gemini Pro 1.0 | Match | +0.2286 | <0.0001 | √ |
| GPT-3.5-turbo vs Gemini Pro 1.0 | Match | +0.1429 | 0.0157 | √ |

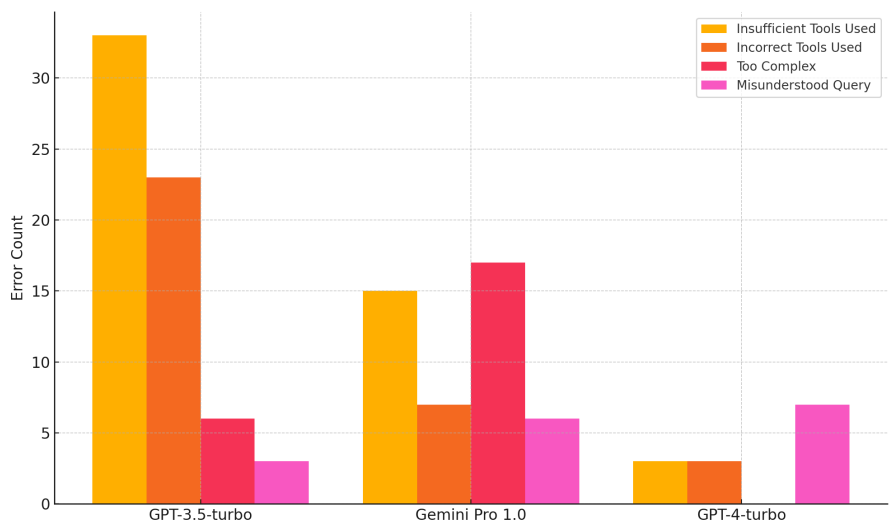

Fig. 5 Error type distribution of three LLM-based agents.

Table 6 Hallucination analysis of vanilla GPT-4-turbo across different question types and subtypes.

| Question Type | Subtype | Vanilla GPT-4-turbo | Hallucination Reason |
|---|---|---|---|
| Whether | / | × | Inaccurate Change Reasoning |
| Size | Basic | √ | / |
|  | Others | × | Inaccurate Change Reasoning |
| Number | Local Area | × | Lack of Local Region Support |
|  | Others | √ | / |
| Class | Local Area | × | Lack of Local Region Support |

| | | |
|---|---|---|
| Others | √ | / |

To further substantiate the claim that our hierarchical agent framework effectively mitigates hallucination issues, we conducted a structured comparison between the proposed agent system and the vanilla GPT-4-turbo model across all question types and subtypes defined in our evaluation dataset. The results, summarized in Table 6, provide quantitative evidence of the hallucination patterns observed in the vanilla GPT-4-turbo setting and highlight the advantages of the ChangeGPT framework.

In Table 6, a checkmark (√) indicates that the vanilla GPT-4-turbo successfully answered all questions within a given category, while a cross (×) denotes complete failure across all corresponding queries. Notably, vanilla GPT-4-turbo was able to reliably answer only those questions in Number (Basic, Comparison) and Class (Whole Image) types. These tasks can be handled using GPT-4's built-in encoder, which allows for basic image understanding and recognition. However, when tasked with questions requiring reasoning over change (e.g., Whether and all subtypes of Size), GPT-4-turbo consistently failed. Although the model sometimes attempted to simulate quantitative analysis (e.g., generating NumPy-based code for pixel difference calculation), it lacked access to specialized remote sensing tools, and often hallucinated results based on incorrect or oversimplified assumptions.

We identified two major sources of hallucination in the vanilla GPT-4-turbo model. The first, Inaccurate Change Reasoning, occurred in tasks such as determining whether a change occurred or estimating change size. In these cases, the model lacked semantic understanding of what constitutes a change in remote sensing context, and failed to apply appropriate thresholding or spatial logic, leading to incorrect or fabricated outputs. The second category, Lack of Local Region Support, emerged in questions requiring spatially localized analysis, such as changes in cropped local regions. Since the vanilla model has no persistent memory or referencing mechanism for localized visual inputs, it could not maintain or resolve distinctions between full-scene images and user-specified subregions, resulting in ambiguous or irrelevant answers.

In fact, the significant performance gap observed between our ChangeGPT and the vanilla GPT-4-turbo baseline, as shown in Table 3 ($\Delta = +0.443$, $p < 0.001$), is largely attributed to the latter's susceptibility to hallucination in complex change analysis tasks. Our agent framework addresses this issue through a combination of design choices. Specifically, the agent incorporates a suite of external visual tools—such as segmentation, detection, and quantitative pixel analysis—to replace uncertain LLM caculation reasoning with grounded, verifiable outputs. The user interface constrains data input operations like cropping, ensuring that all visual content is accurately framed and semantically disambiguated before reasoning begins. In parallel, the system maintains a persistent image reference mechanism that keeps track of image identities across multiple reasoning steps, allowing the agent to accurately perform localized analysis, something that vanilla LLMs typically struggle with due to the lack of spatial grounding. Taken together, these measures form a coherent mitigation strategy that not only prevents typical hallucination modes, such as unjustified numeric claims or spatial mismatches, but also enhances interpretability, reproducibility, and overall answer correctness across diverse remote sensing change analysis scenarios.

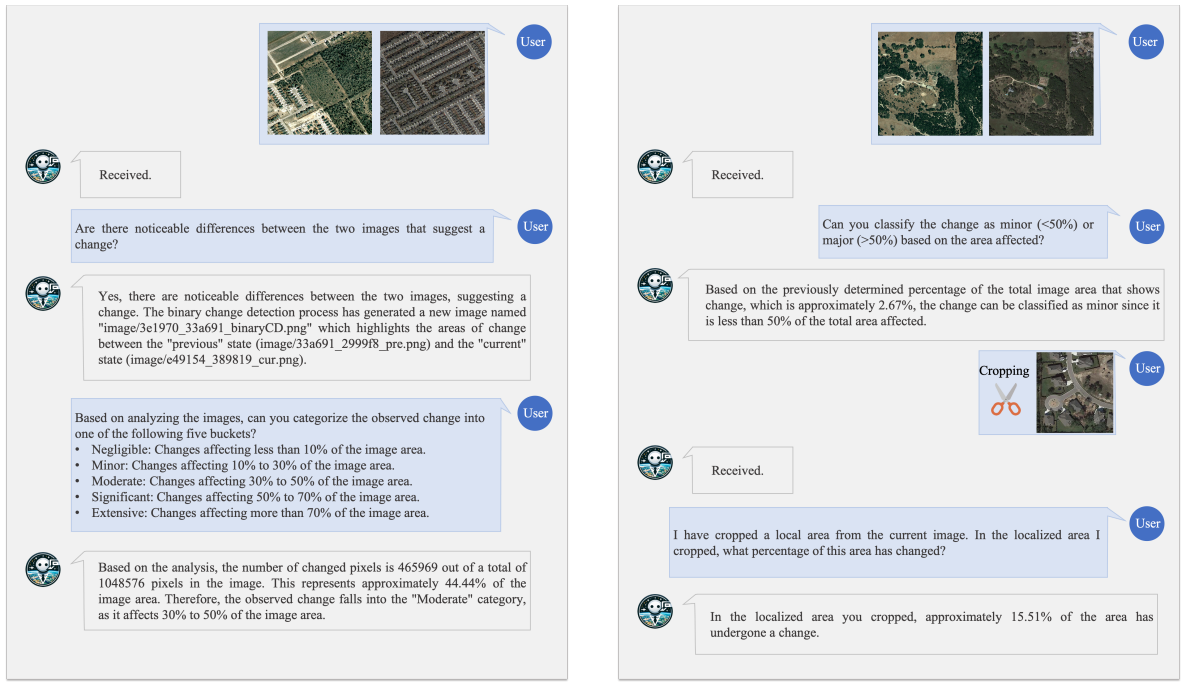

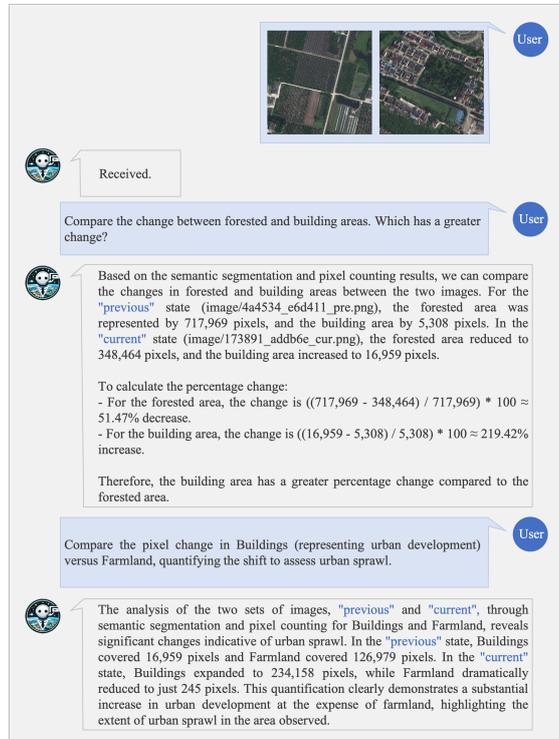

Fig. 6 Examples of multi-round dialogues with ChangeGPT agent.

To better present the workflow of ChangeGPT, three examples are selected in Fig 6. In Fig. 6(a), we tested questions of both 'Whether' and 'Size' types. The agent responded with the correct answers and demonstrated its reasoning ability for the bucketing requirement. In Fig. 6(b), users provided not only language queries but also direct human interaction. Specifically, a local area of the current image was cropped and queried with questions. The agent successfully distinguished between the whole image (previous or current) and the cropped area, verifying its image sensitivity ability. This example further highlights the robustness and adaptability of our agent in handling diverse and dynamic queries in practical application scenarios. Fig. 6(c) presents the ability of the agent to manage a complex task involving both calculation and comparison to generate a final analytical report. In this example, the agent effectively planned a sequence of subtasks and selected the appropriate tools for each subtask, demonstrating its planning capabilities. These examples collectively underscore the versatility and capability of ChangeGPT in addressing various queries about analysing changes in remote sensing applications.

**4.3. Case Study**

In addition to experimental evaluation, we conducted a case study to verify our agent's effectiveness in real-world scenarios, in Qianhai Bay, Shenzhen City, China, highlighting its capabilities in monitoring urban changes and supporting sustainable urban design. Qianhai Bay is part of the Qianhai Shenzhen-Hong Kong Modern Service Industry Cooperation Zone, encompassing a total area of 120.56 square kilometers, which plays a crucial role in enhancing cooperation between Guangdong, Hong Kong, and Macao. Since its establishment, Qianhai Bay has experienced rapid and significant development. Assessing and analysing the changes in this zone over the past decades is vital for supporting sustainable design initiatives, including environmental monitoring, land use planning, and social development.

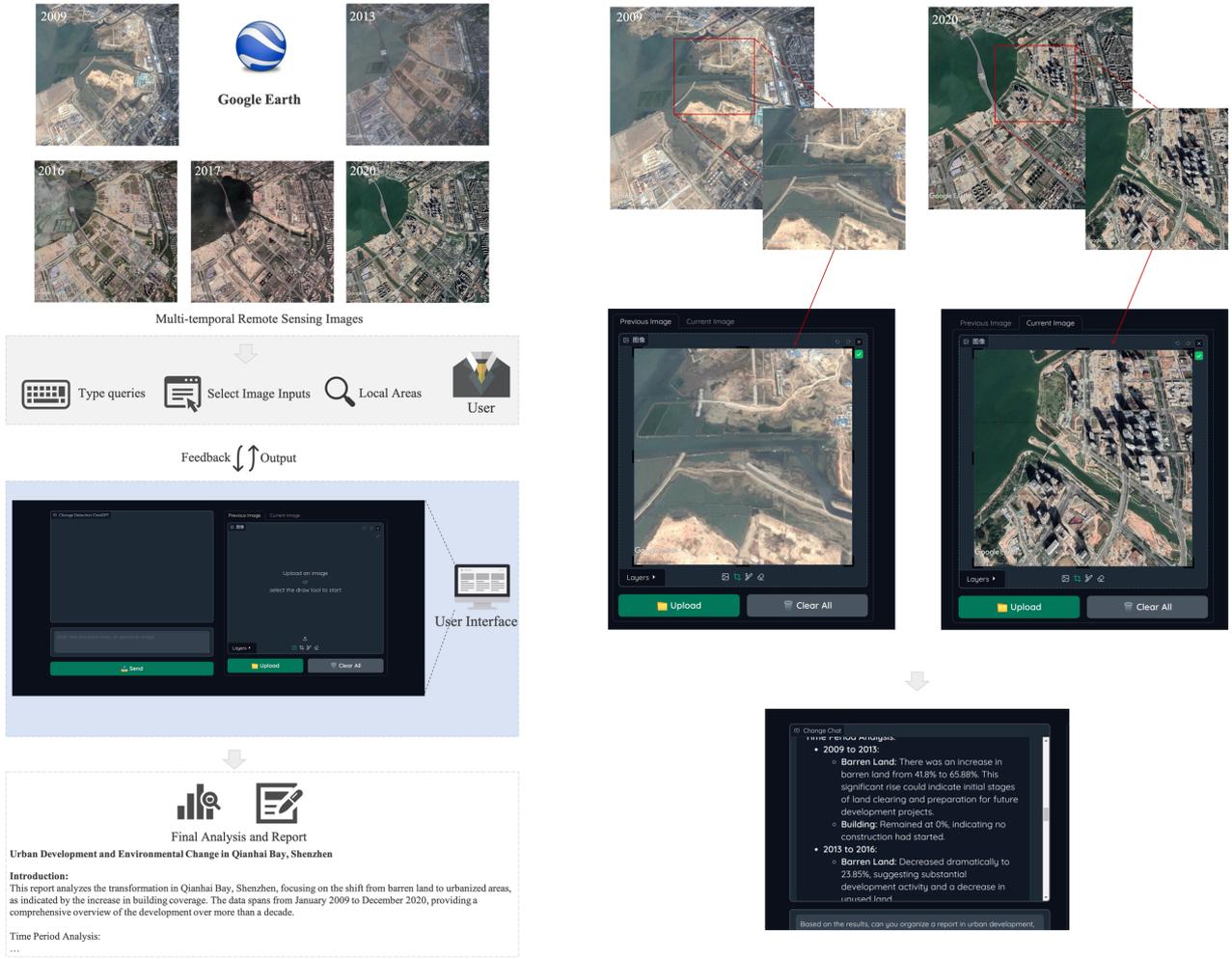

*Fig. 7 Overview workflow of case study.*

As shown in Fig 7, we firstly collect multiple temporal remote sensing images of Qianhai Bay from Google Earth. We then cropped two distinct areas, one near the sea and the other near the center of Nanshan district, to examine changes from various perspectives. Utilizing these areas, we process multi-round dialogues with our agents through a designed user interface to understand the diverse aspects of changes in Qianhai Bay. Subsequently, leveraging the acquired data on changes, we query an analytical report for sustainable design. Changes in the size of different land use classes play an important part in the analytical report, which we assessed through semantic segmentation models. To this end, we employed three different models: HRNet [54], UNetFormer [55], and DCSwin. These models were trained on the LoveDA remote sensing semantic segmentation dataset, and subsequently, our research team manually labeled the cropped areas in Qianhai Bay for evaluation. Following the setup of the LoveDA dataset, our segmentation masks were annotated with seven classes: background, water, barren, road, building, forest, and farmland.

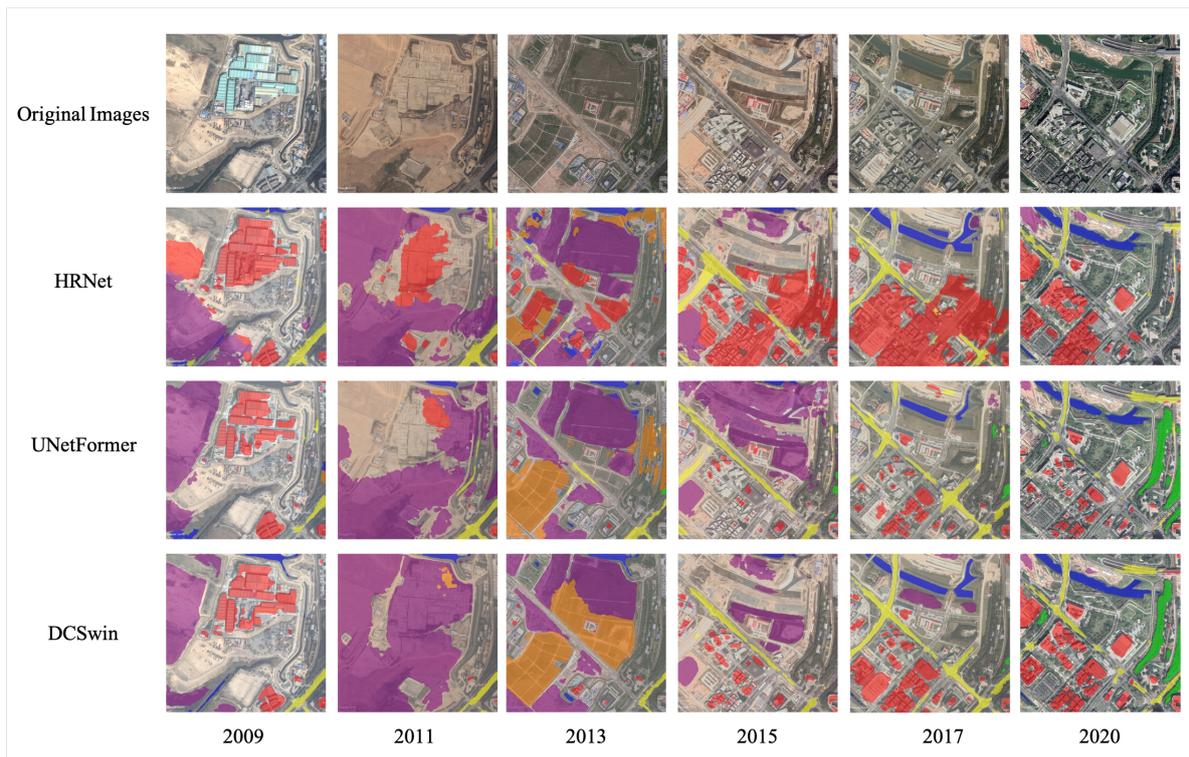

Fig. 8 Segmentation results visualization of different models in the first cropped area near the center of Nanshan district in Qianhai Bay.

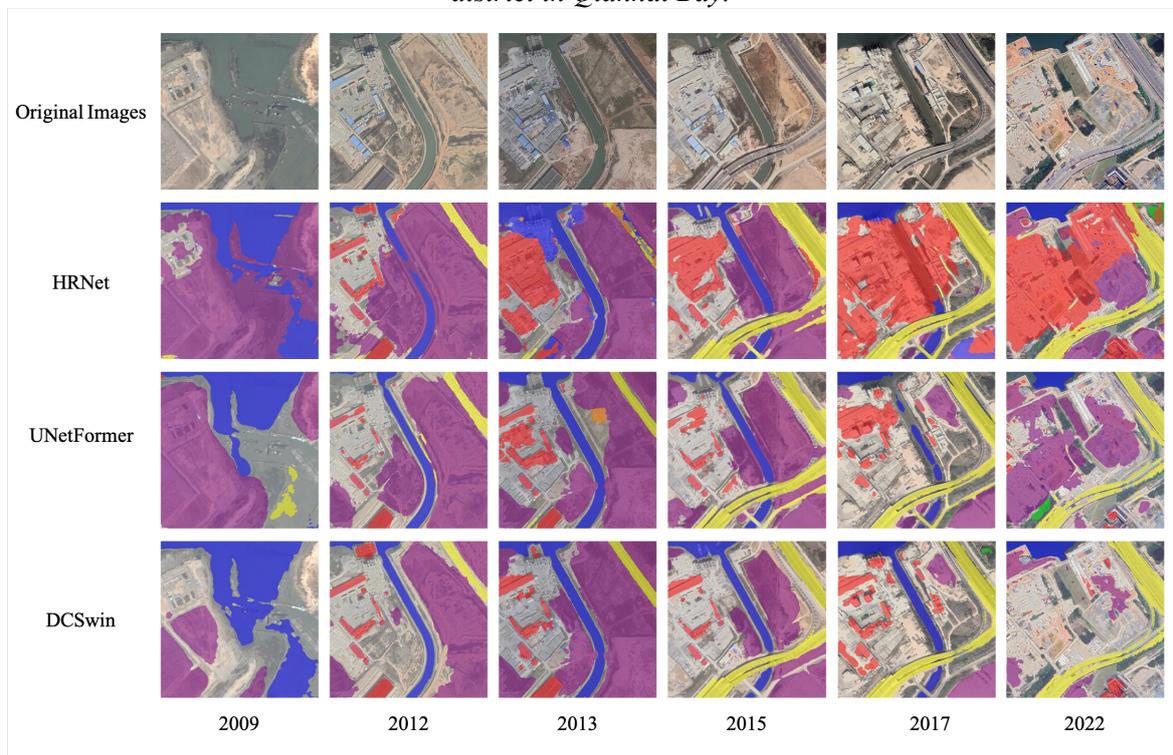

Fig. 9 Segmentation results visualization of different models in the second cropped area near the sea in Qianhai Bay.

Table 7. Segmentation performance of different models on cropped areas:
(a) First area is located near the center of Nanshan District.
(b) Second area is situated near the coast.

| Model | HRNet | UNetFormer | DCSwin |
|---|---|---|---|
| Background | 66.01 | 75.27 | 79.43 |
| Water | 70.52 | 48.46 | 73.16 |
| Barren | 42.22 | 49.56 | 64.89 |
| Road | 34.63 | 38.69 | 50.23 |
| Building | 74.23 | 33.54 | 52.94 |
| Forest | 0 | 74.33 | 73.98 |
| Farmland | 16.91 | 61.84 | 97.81 |
| **OA** | 59.40 | 63.70 | 73.06 |
| **mIoU** | 29.86 | 40.31 | 55.10 |
| **Mean F1** | 46.63 | 39.38 | 53.56 |

| Model | HRNet | UNetFormer | DCSwin |
|---|---|---|---|
| Background | 32.87 | 61.33 | 82.23 |
| Water | 69.11 | 55.10 | 76.88 |
| Barren | 88.46 | 91.84 | 80.89 |
| Road | 80.61 | 79.26 | 85.82 |
| Building | 91.33 | 60.08 | 69.81 |
| Forest* | / | / | / |
| Farmland* | / | / | / |
| **OA** | 54.75 | 68.38 | 81.07 |
| **mIoU** | 41.00 | 49.77 | 63.93 |
| **Mean F1** | 48.34 | 54.43 | 77.27 |

Segmentation results of different models on our cropped areas (the first one near the center of Nanshan district and the second one near the sea) are visualized in Fig 8 and Fig 9. Table 7 presents the he comparative results of different models for the cropped areas, and it can be seen that the DCSwin performs better than the other two models. In Fig 8, the parks are not well segmented due to the fact that the model is trained with LoveDA dataset and the parks are not within the training labels. Low vegetation covered areas like parks are neither forest nor farmland, so they cannot be well segmented.

Apart from these three general models trained with LoveDA dataset, we also tried SAMGEO [56,57], which is a newly developed remote sensing segmentation model adapted from Segment Anything Model (SAM) [58] for geospatial data. The results are visualized in Fig 10. As shown in Fig 10, it is capable of generating more detailed segmentation masks, focusing on instance segmentation rather than semantic segmentation, and the labels are class-agnostic. This setting currently does not align with the requirements of our agent. However, as it evolves and offers more functionalities in the future, we may consider adopting it into our agent.

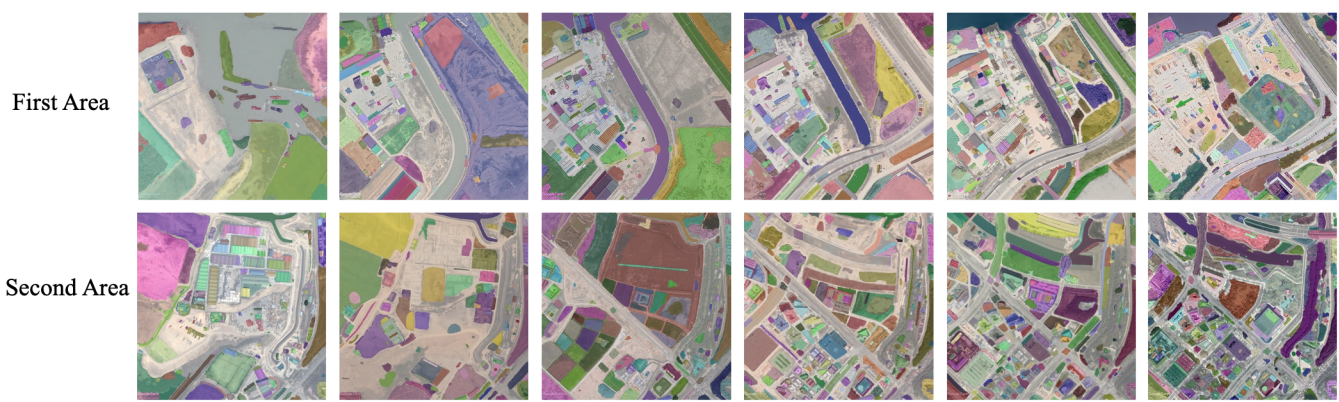

Fig. 10 Segmentation results of SAMGEO model, which only generates instance segmentation results.

Based on the results of change analysis from our agent, several notable trends and patterns can be observed. There is a dramatic decrease in barren land, from 59.11% in 2011 to 0% in 2020 while the percentage of roads has increased steadily from 1.65% in 2009 to 10.53% in 2020. Building coverage shows variability, peaking at 16% in 2009, decreasing to 0% in 2011, and then stabilizing around 14% from 2015 to 2020. The dramatic decrease in barren land from 59.11% in 2011 to 0% in 2020, coupled with the substantial increase in developed land (roads and buildings), suggests that land is being rapidly converted for development purposes. This could imply overdevelopment, potentially leading to environmental degradation and loss of natural habitats.

## 4.4. Discussion

This subsection provides an in-depth discussion of key aspects of the proposed framework, including its generalizability, computational efficiency, interpretability, and ethical considerations. These reflections aim to contextualize the practical implications and guide future applications and development.

### 4.4.1. Generalization across Diverse Geographies and Urban Morphologies

While the case study in Section 4.3 focuses on Qianhai Bay, the selected subregions within this area, specifically a dense inland district and a coastal zone, exhibit significantly different spatial patterns and urban morphologies. This intra-city variation provides a valuable preliminary lens to assess the adaptability of the ChangeGPT framework across differing urban forms, even within a single geographic region.

We observed that while the reasoning structure of the agent remains stable, the performance of underlying visual tools (e.g., segmentation models) varies depending on the land use characteristics and their representation in the training dataset. This highlights a limitation in tool model transferability rather than in the agent architecture itself. Despite these differences, ChangeGPT was still able to extract meaningful change patterns and synthesize relevant insights across the two areas, indicating that its reasoning process remains robust even when visual input quality is imperfect or context-sensitive.

This observation highlights that the generalization capability of ChangeGPT does not rely on a single pretrained model performing uniformly well across all environments, but rather stems from its modular and flexible framework design. When faced with varied urban morphologies or climatic conditions, the backend tools, particularly vision models such as segmentation or detection modules, can be selectively replaced, retrained, or fine-tuned on domain-specific datasets to better match the spatial and semantic characteristics of a given region. In this way, ChangeGPT can maintain performance consistency and analytical reliability even in regions that differ significantly from the original training distribution. As more geographically diverse and fine-grained datasets become available, the framework is well-positioned to support the integration of context-specific models. The agent's use of structured prompts and natural language reasoning further ensures that once appropriately adapted tools are in place, the overall workflow remains robust, interpretable, and transferable to a wide range of geospatial scenarios.

Although the current case study does not span multiple cities or climate zones, it reveals promising evidence that the framework's design supports transferability and adaptation. Future work will explore its application across broader geographies, such as high-density inland metropolises, arid environments, and rapidly urbanizing peri-urban regions, to further assess and enhance its generalization capabilities.

### 4.4.2. Computational Cost, Scalability, and Optimization Strategies for Real-World Applications

The practical deployment of the ChangeGPT framework involves several important considerations related to computational cost, scalability, and inference latency, especially in the context of real-time or large-scale remote sensing applications. These performance factors are influenced primarily by two components of the framework: the tool models and the LLM backend.

On the tool level, each visual component integrated into ChangeGPT, such as segmentation, detection, or classification models, incurs its own computational footprint. This cost is largely determined by the model's architecture, parameter size, and inference time. For example, lightweight models such as YOLOv5 or MobileNet-based classifiers are suitable for time-sensitive or resource-constrained environments, offering rapid responses at the expense of some accuracy. In contrast, high-capacity models deliver greater precision but demand significantly more memory and processing time. Depending on the requirements of a specific application, users or system designers can substitute models to strike a balance between accuracy and efficiency.

A more substantial computational consideration lies in the LLM backend, which governs the agent's reasoning and coordination capabilities. In deployment contexts, one key decision is whether to use on-premise LLM deployment or cloud-based inference via APIs. Each option entails distinct trade-offs. Local deployment offers the benefits of lower latency, increased control, and improved data privacy. However, hosting LLMs locally (especially larger models) demands considerable computing infrastructure, as well as ongoing maintenance and update overhead. By contrast, using cloud-based APIs provides access to state-of-the-art reasoning capabilities without the need for local infrastructure. This can enable faster iteration and superior accuracy for complex, multi-step queries. Nevertheless, it introduces external dependencies, potential network latency, usage-based costs, and data confidentiality risks, which must be carefully assessed for operational deployments.

In terms of scalability, ChangeGPT's modular agent design naturally supports deployment in distributed and microservice-based environments. The framework separates image analysis modules, the LLM interface, and the reasoning controller into distinct components that can run independently and scale horizontally. In multi-user or high-throughput scenarios, performance can be further enhanced through techniques such as concurrent tool invocation, output caching, and adaptive task scheduling.

Table 8. Estimated computational cost under varying query complexity levels, including tool usage and LLM coordination latency.

| Difficulty Level | Tool Count | Typical Tool Types | API Rounds(≈ Tools + 2) | Avg. Tool Inference Time(ms) | Avg. API Time per Call(s) | Estimated Total Latency(s/query) |
|---|---|---|---|---|---|---|
| **Easy** | 1 | Binary Change Detection (SAM & CLIP), Scene Classification (ResNet) | 3 | 150–500 | 2.5–5.0 | ~3.0–7.0 |
| **Medium** | 2 | Object Detection (YOLOv5), Semantic Segmentation (DCSwin) | 4 | 200–600 | 2.5–5.0 | ~7.0–13.0 |
| **Difficult** | >2 (typically 4, up to 8+) | Object Detection (YOLOv5), Semantic Segmentation (DCSwin), multi-round Pixel Counting | 6–10 | 400–1500+ | 2.5–5.0 | ~15.0–35.0+ |

To provide a clearer understanding of how task complexity translates into computational overhead, we present a latency estimation based on the number of tools required per query and the corresponding LLM coordination steps. As described in Section 4.2, questions in our question dataset are categorized into Easy, Medium, and Difficult levels according to the number of tools involved. Table 8 summarizes the expected inference time for each level of complexity, considering typical visual model latency, the number of LLM API rounds (which is approximately equal to the number of tools plus two), and the average system response time per API call.

These estimations indicate that Easy queries can often be completed within 3 to 7 seconds, while Medium and Difficult queries may range from 7 to over 30 seconds. Such latency profiles underscore the importance of optimization strategies like local LLM hosting, or model compression in real-world deployment scenarios.

In summary, although the current implementation of ChangeGPT is designed primarily as a research prototype, its architectural flexibility and modular design make it highly compatible with a range of practical optimization strategies. These characteristics lay a strong foundation for its deployment in real-world operational contexts, particularly when computational resources and response times are critical constraints.

### 4.4.3. Interpretability and Explainability

Interpretability and explainability are increasingly critical in remote sensing change analysis, especially when the results are expected to support decision-making in urban planning, environmental monitoring, and disaster response. Unlike conventional change detection models that typically offer only binary or semantic change maps, our agent-based framework, ChangeGPT, offers a fundamentally different approach to explainability by explicitly modeling and exposing the reasoning process through its hierarchical structure.

One of the key components supporting explainability in our system is the Reference Layer within the Planning Navigator module, as described in Section 3.2. This layer plays a pivotal role in recording, retrieving, and leveraging historical context, including past user queries, reasoning steps, tool invocations, and intermediate outputs. By retaining this history, the agent not only avoids redundant operations but also provides transparency into the decision-making process. For instance, if a user revisits a past query or questions a particular result, the system can trace back through the reference data to reveal which tools were selected, what intermediate results were generated, and how each step contributed to the final answer.

Furthermore, this architecture allows for future extensibility. The Reference Layer can be expanded to support longer history windows, richer storage formats, and even real-time memory updates. This would enhance the system's interpretability by providing a complete and persistent trace of analytical logic. Additionally, the tool-based modular structure itself is inherently interpretable: since each tool in the execution plan is functionally isolated and explicitly invoked, users and developers can inspect the behavior of each module in the reasoning chain.

Overall, ChangeGPT supports interpretability both through structural transparency (in how queries are decomposed and processed) and through historical traceability (in how actions and decisions are recorded). This offers a strong foundation for trust and auditability, distinguishing it from traditional black-box models. We believe this design aligns well with the growing demand for interpretable AI in remote sensing applications.

### 4.4.4. Ethical Considerations and Risk Awareness

As the adoption of LLMs and multi-modal AI systems continues to expand in the field of urban analysis, it is essential to address their potential ethical and societal implications. In the context of ChangeGPT, two key areas of concern are the inherent biases of LLMs and the risk of misinterpretation in remote sensing analysis. First, LLMs are known to reflect biases present in their pretraining data, which may include geographic, socioeconomic, and cultural skew. In urban applications, this may manifest in uneven performance across regions, underrepresentation of certain urban features, or assumptions that do not hold in diverse real-world contexts. While ChangeGPT mitigates this risk through tool-based visual analysis and explicit spatial reasoning, the language model still plays a central role in query interpretation and task planning. Therefore, users should be cautious when applying the system across different regions and ensure that the results are validated against ground-truth knowledge or expert review, especially in policy-sensitive domains such as urban planning or disaster response.

Second, remote sensing imagery inherently carries interpretation ambiguity. Factors such as atmospheric conditions, image resolution, temporal gaps, or sensor variability can affect the perceived nature of changes on the ground. When mediated through an AI system, particularly one involving multi-step reasoning, the risk of misinterpretation can be compounded. ChangeGPT addresses this issue in part by grounding its outputs in tool-generated visual evidence and encouraging step-by-step task decomposition. Nevertheless, the possibility of misidentifying change types or misjudging the significance of detected variations remains, particularly when operating in unfamiliar or data-scarce areas.

More broadly, the societal implications of deploying AI-driven decision-making systems must be carefully considered. Systems like ChangeGPT should not replace expert judgment or local contextual knowledge, but rather serve as analytical aids. Transparency in how outputs are generated, reproducibility of reasoning processes, and the ability to audit intermediate steps are vital to building user trust and ensuring responsible use. Future work should continue to investigate fairness, accountability, and human-in-the-loop governance mechanisms for AI in urban and environmental applications.

## 5. Limitation

While the ChangeGPT framework demonstrates adaptability and generalization across different urban environments, certain limitations related to the tool generalization have been identified. Specifically, in the case study conducted in Qianhai Bay, we observed that the performance of underlying visual tools, particularly segmentation models, was suboptimal in certain regions, such as parks and low-vegetation zones. This limitation arose from the insufficient representation of these land cover types in the training data, which led to poor segmentation results in those areas.

However, this limitation also provides a path for future improvement. ChangeGPT's modular design allows for the flexible replacement, retraining, or fine-tuning of the backend tools, which means that the framework can be adapted to specific tasks and environments. For instance, in real-world applications, a more targeted training process can be employed to better represent the features of interest, ensuring that tools are more capable of handling specific urban contexts and land cover types. Additionally, selecting tools that are particularly suited to the task at hand, whether through further training or by integrating more specialized models, can improve overall performance and ensure reliable results.

Thus, while the case study highlights the potential limitations in tool generalization, it should be noted that these issues can be mitigated in practice by ensuring that the tools used are well-suited and sufficiently trained for the specific application scenario. In future work, we plan to address these limitations by incorporating more diverse datasets and developing more specialized models to better handle the challenges of real-world urban change analysis.

Another limitation concerns the lack of external ground-truth validation in the Qianhai Bay case study. Due to the historical nature of the satellite imagery used, it is inherently difficult to obtain corresponding ground-truth information through on-site validation, as it is not feasible to retrospectively access the conditions present at the time the images were captured. Additional constraints related to cost and accessibility further limit the possibility of third-party data collection. As an alternative, our evaluation relied on expert annotations created by our team, using the semantic class structure defined in the LoveDA dataset as a consistent labeling reference. We acknowledge that future work would benefit from collaborating with local authorities to enhance the robustness of the evaluation.

## 6. Conclusion

This paper presented a general agent framework for intelligent change assessment and analysis in remote sensing images. Targeting practical application scenarios, the developed ChangeGPT aims to address the limitations of current change detection models, which often fail to support diverse types of changes requiring subsequent analysis. Specifically, a hierarchical framework was designed to decompose queries, plan intermediate reasoning steps, and invoke appropriate vision tools to accomplish complex change analysis tasks. A specialized toolkit was constructed to integrate multiple remote sensing Vision Foundation Models (VFMs), and a hallucination mitigation mechanism was introduced to improve reliability. To evaluate the effectiveness, a 140-question dataset categorized by real-world scenarios was constructed and task-specific metrics were defined, including Precision, Recall, and Match. Experimental results show that the agent powered by GPT-4-turbo achieved a Match rate of 90.71%, significantly outperforming variants based on GPT-3.5-turbo (53.57%) and Gemini Pro 1.0 (68.57%). Additionally, we conducted a case study in Qianhai Bay, Shenzhen, China, where the agent successfully responded to multi-step analytical queries involving land use change, area estimation, and urban development analysis. These results validate the framework's practical effectiveness and robustness in handling real-world urban change analysis. In future work, we aim to enrich the system by incorporating domain-specific tools (e.g., SAR processing modules, land-use simulation models) and enhance adaptability across diverse geographic and temporal settings, with the goal of supporting broader applications in urban planning, environmental monitoring, and geospatial decision-making.

**Declaration of Generative AI and AI-assisted technologies in the writing process**

Statement: During the preparation of this work the author(s) used ChatGPT-4 in order to improve readability and language. After using this tool/service, the author(s) reviewed and edited the content as needed and take(s) full responsibility for the content of the publication.

Katariya, S. Riedel, P. Bailey, K. Xiao, N. Ghelani, L. Aroyo, A. Slone, N. Houlsby, X. Xiong, Z. Yang, E. Gribovskaya, J. Adler, M. Wirth, L. Lee, M. Li, T. Kagohara, J. Pavagadhi, S. Bridgers, A. Bortsova, S. Ghemawat, Z. Ahmed, T. Liu, R. Powell, V. Bolina, M. Iinuma, P. Zablotskaia, J. Besley, D.-W. Chung, T. Dozat, R. Comanescu, X. Si, J. Greer, G. Su, M. Polacek, R.L. Kaufman, S. Tokumine, H. Hu, E. Buchatskaya, Y. Miao, M. Elhawaty, A. Siddhant, N. Tomasev, J. Xing, C. Greer, H. Miller, S. Ashraf, A. Roy, Z. Zhang, A. Ma, A. Filos, M. Besta, R. Blevins, T. Klimenko, C.-K. Yeh, S. Changpinyo, J. Mu, O. Chang, M. Pajarskas, C. Muir, V. Cohen, C.L. Lan, K. Haridasan, A. Marathe, S. Hansen, S. Douglas, R. Samuel, M. Wang, S. Austin, C. Lan, J. Jiang, J. Chiu, J.A. Lorenzo, L.L. Sjösund, S. Cevey, Z. Gleicher, T. Avrahami, A. Boral, H. Srinivasan, V. Selo, R. May, K. Aisopos, L. Hussenot, L.B. Soares, K. Baumli, M.B. Chang, A. Recasens, B. Caine, A. Pritzel, F. Pavetic, F. Pardo, A. Gergely, J. Frye, V. Ramasesh, D. Horgan, K. Badola, N. Kassner, S. Roy, E. Dyer, V.C. Campos, A. Tomala, Y. Tang, D.E. Badawy, E. White, B. Mustafa, O. Lang, A. Jindal, S. Vikram, Z. Gong, S. Caelles, R. Hemsley, G. Thornton, F. Feng, W. Stokowiec, C. Zheng, P. Thacker, Ç. Ünlü, Z. Zhang, M. Saleh, J. Svensson, M. Bileschi, P. Patil, A. Anand, R. Ring, K. Tsihlas, A. Vezer, M. Selvi, T. Shevlane, M. Rodriguez, T. Kwiatkowski, S. Daruki, K. Rong, A. Dafoe, N. FitzGerald, K. Gu-Lemberg, M. Khan, L.A. Hendricks, M. Pellat, V. Feinberg, J. Cobon-Kerr, T. Sainath, M. Rauh, S.H. Hashemi, R. Ives, Y. Hasson, E. Noland, Y. Cao, N. Byrd, L. Hou, Q. Wang, T. Sottiaux, M. Paganini, J.-B. Lespiau, A. Moufarek, S. Hassan, K. Shivakumar, J. van Amersfoort, A. Mandhane, P. Joshi, A. Goyal, M. Tung, A. Brock, H. Sheahan, V. Misra, C. Li, N. Rakićević, M. Dehghani, F. Liu, S. Mittal, J. Oh, S. Noury, E. Sezener, F. Huot, M. Lamm, N. De Cao, C. Chen, S. Mudgal, R. Stella, K. Brooks, G. Vasudevan, C. Liu, M. Chain, N. Melinkeri, A. Cohen, V. Wang, K. Seymore, S. Zubkov, R. Goel, S. Yue, S. Krishnakumaran, B. Albert, N. Hurley, M. Sano, A. Mohananey, J. Joughin, E. Filonov, T. Kępa, Y. Eldawy, J. Lim, R. Rishi, S. Badiezadegan, T. Bos, J. Chang, S. Jain, S.G.S. Padmanabhan, S. Puttagunta, K. Krishna, L. Baker, N. Kalb, V. Bedapudi, A. Kurzrok, S. Lei, A. Yu, O. Litvin, X. Zhou, Z. Wu, S. Sobell, A. Siciliano, A. Papir, R. Neale, J. Bragagnolo, T. Toor, T. Chen, V. Anklin, F. Wang, R. Feng, M. Gholami, K. Ling, L. Liu, J. Walter, H. Moghaddam, A. Kishore, J. Adamek, T. Mercado, J. Mallinson, S. Wandekar, S. Cagle, E. Ofek, G. Garrido, C. Lombriser, M. Mukha, B. Sun, H.R. Mohammad, J. Matak, Y. Qian, V. Peswani, P. Janus, Q. Yuan, L. Schelin, O. David, A. Garg, Y. He, O. Duzhyi, A. Älgmyr, T. Lottaz, Q. Li, V. Yadav, L. Xu, A. Chinien, R. Shivanna, A. Chuklin, J. Li, C. Spadine, T. Wolfe, K. Mohamed, S. Das, Z. Dai, K. He, D. von Dincklage, S. Upadhyay, A. Maurya, L. Chi, S. Krause, K. Salama, P.G. Rabinovitch, P.K.R. M, A. Selvan, M. Dektiarev, G. Ghiasi, E. Guven, H. Gupta, B. Liu, D. Sharma, I.H. Shtacher, S. Paul, O. Akerlund, F.-X. Aubet, T. Huang, C. Zhu, E. Zhu, E. Teixeira, M. Fritze, F. Bertolini, L.-E. Marinescu, M. Bölle, D. Paulus, K. Gupta, T. Latkar, M. Chang, J. Sanders, R. Wilson, X. Wu, Y.-X. Tan, L.N. Thiet, T. Doshi, S. Lall, S. Mishra, W. Chen, T. Luong, S. Benjamin, J. Lee, E. Andrejczuk, D. Rabiej, V. Ranjan, K. Styrc, P. Yin, J. Simon, M.R. Harriott, M. Bansal, A. Robsky, G. Bacon, D. Greene, D. Mirylenka, C. Zhou, O. Sarvana, A. Goyal, S. Andermatt, P. Siegler, B. Horn, A. Israel, F. Pongetti, C.-W. "Louis" Chen, M. Selvatici, P. Silva, K. Wang, J. Tolins, K. Guu, R. Yogev, X. Cai, A. Agostini, M. Shah, H. Nguyen, N.Ó. Donnaile, S. Pereira, L. Friso, A. Stambler, A. Kurzrok, C. Kuang, Y. Romanikhin, M. Geller, Z. Yan, K. Jang, C.-C. Lee, W. Fica, E. Malmi, Q. Tan, D. Banica, D. Balle, R. Pham, Y. Huang, D. Avram, H. Shi, J. Singh, C. Hidey, N. Ahuja, P. Saxena, D. Dooley, S.P. Potharaju, E. O'Neill, A. Gokulchandran, R. Foley, K. Zhao, M. Dusenberry, Y. Liu, P. Mehta, R. Kotikalapudi, C. Safranek-Shrader, A. Goodman, J. Kessinger, E. Globen, P. Kolhar, C. Gorgolewski, A. Ibrahim, Y. Song, A. Eichenbaum, T. Brovelli, S. Potluri, P. Lahoti, C. Baetu, A. Ghorbani, C. Chen, A. Crawford, S. Pal, M. Sridhar, P. Gurita, A. Mujika, I. Petrovski, P.-L. Cedoz, C. Li, S. Chen, N.D. Santo, S. Goyal, J. Punjabi, K. Kappaganthu, C. Kwak, P. LV, S. Velury, H. Choudhury, J. Hall, P. Shah, R. Figueira, M. Thomas, M. Lu, T. Zhou, C. Kumar, T. Jurdi, S. Chikkerur, Y. Ma, A. Yu, S. Kwak, V. Ähdel, S. Rajayogam, T. Choma, F. Liu, A. Barua, C. Ji, J.H. Park, V. Hellendoorn, A. Bailey, T. Bilal, H. Zhou, M. Khatir, C. Sutton, W. Rzadkowski, F. Macintosh, K. Shagin, P. Medina, C. Liang, J. Zhou, P. Shah, Y. Bi, A. Dankovics, S. Banga, S. Lehmann, M. Bredesen, Z. Lin, J.E.
47

Hoffmann, J. Lai, R. Chung, K. Yang, N. Balani, A. Bražinskas, A. Sozanschi, M. Hayes, H.F. Alcalde, P. Makarov, W. Chen, A. Stella, L. Snijders, M. Mandl, A. Kärrman, P. Nowak, X. Wu, A. Dyck, K. Vaidyanathan, R. R, J. Mallet, M. Rudominer, E. Johnston, S. Mittal, A. Udathu, J. Christensen, V. Verma, Z. Irving, A. Santucci, G. Elsayed, E. Davoodi, M. Georgiev, I. Tenney, N. Hua, G. Cideron, E. Leurent, M. Alnahlawi, I. Georgescu, N. Wei, I. Zheng, D. Scandinaro, H. Jiang, J. Snoek, M. Sundararajan, X. Wang, Z. Ontiveros, I. Karo, J. Cole, V. Rajashekhar, L. Tumeh, E. Ben-David, R. Jain, J. Uesato, R. Datta, O. Bunyan, S. Wu, J. Zhang, P. Stanczyk, Y. Zhang, D. Steiner, S. Naskar, M. Azzam, M. Johnson, A. Paszke, C.-C. Chiu, J.S. Elias, A. Mohiuddin, F. Muhammad, J. Miao, A. Lee, N. Vieillard, J. Park, J. Zhang, J. Stanway, D. Garmon, A. Karmarkar, Z. Dong, J. Lee, A. Kumar, L. Zhou, J. Evens, W. Isaac, G. Irving, E. Loper, M. Fink, I. Arkatkar, N. Chen, I. Shafran, I. Petrychenko, Z. Chen, J. Jia, A. Levskaya, Z. Zhu, P. Grabowski, Y. Mao, A. Magni, K. Yao, J. Snaider, N. Casagrande, E. Palmer, P. Suganthan, A. Castaño, I. Giannoumis, W. Kim, M. Rybiński, A. Sreevatsa, J. Prendki, D. Soergel, A. Goedeckemeyer, W. Gierke, M. Jafari, M. Gaba, J. Wiesner, D.G. Wright, Y. Wei, H. Vashisht, Y. Kulizhskaya, J. Hoover, M. Le, L. Li, C. Iwuanyanwu, L. Liu, K. Ramirez, A. Khorlin, A. Cui, T. LIN, M. Wu, R. Aguilar, K. Pallo, A. Chakladar, G. Perng, E.A. Abellan, M. Zhang, I. Dasgupta, N. Kushman, I. Penchev, A. Repina, X. Wu, T. van der Weide, P. Ponnapalli, C. Kaplan, J. Simsa, S. Li, O. Dousse, F. Yang, J. Piper, N. Ie, R. Pasumarthi, N. Lintz, A. Vijayakumar, D. Andor, P. Valenzuela, M. Lui, C. Paduraru, D. Peng, K. Lee, S. Zhang, S. Greene, D.D. Nguyen, P. Kurylowicz, C. Hardin, L. Dixon, L. Janzer, K. Choo, Z. Feng, B. Zhang, A. Singhal, D. Du, D. McKinnon, N. Antropova, T. Bolukbasi, O. Keller, D. Reid, D. Finchelstein, M.A. Raad, R. Crocker, P. Hawkins, R. Dadashi, C. Gaffney, K. Franko, A. Bulanova, R. Leblond, S. Chung, H. Askham, L.C. Cobo, K. Xu, F. Fischer, J. Xu, C. Sorokin, C. Alberti, C.-C. Lin, C. Evans, A. Dimitriev, H. Forbes, D. Banarse, Z. Tung, M. Omernick, C. Bishop, R. Sterneck, R. Jain, J. Xia, E. Amid, F. Piccinno, X. Wang, P. Banzal, D.J. Mankowitz, A. Polozov, V. Krakovna, S. Brown, M. Bateni, D. Duan, V. Firoiu, M. Thotakuri, T. Natan, M. Geist, S. tan Girgin, H. Li, J. Ye, O. Roval, R. Tojo, M. Kwong, J. Lee-Thorp, C. Yew, D. Sinopalnikov, S. Ramos, J. Mellor, A. Sharma, K. Wu, D. Miller, N. Sonnerat, D. Vnukov, R. Greig, J. Beattie, E. Caveness, L. Bai, J. Eisenschlos, A. Korchemniy, T. Tsai, M. Jasarevic, W. Kong, P. Dao, Z. Zheng, F. Liu, F. Yang, R. Zhu, T.H. Teh, J. Sanmiya, E. Gladchenko, N. Trdin, D. Toyama, E. Rosen, S. Tavakkol, L. Xue, C. Elkind, O. Woodman, J. Carpenter, G. Papamakarios, R. Kemp, S. Kafle, T. Grunina, R. Sinha, A. Talbert, D. Wu, D. Owusu-Afriyie, C. Du, C. Thornton, J. Pont-Tuset, P. Narayana, J. Li, S. Fatehi, J. Wieting, O. Ajmeri, B. Uria, Y. Ko, L. Knight, A. Héliou, N. Niu, S. Gu, C. Pang, Y. Li, N. Levine, A. Stolovich, R. Santamaria-Fernandez, S. Goenka, W. Yustalim, R. Strudel, A. Elqursh, C. Deck, H. Lee, Z. Li, K. Levin, R. Hoffmann, D. Holtmann-Rice, O. Bachem, S. Arora, C. Koh, S.H. Yeganeh, S. Põder, M. Tariq, Y. Sun, L. Ionita, M. Seyedhosseini, P. Tafti, Z. Liu, A. Gulati, J. Liu, X. Ye, B. Chrzaszcz, L. Wang, N. Sethi, T. Li, B. Brown, S. Singh, W. Fan, A. Parisi, J. Stanton, V. Koverkathu, C.A. Choquette-Choo, Y. Li, T. Lu, A. Ittycheriah, P. Shroff, M. Varadarajan, S. Bahargam, R. Willoughby, D. Gaddy, G. Desjardins, M. Cornero, B. Robenek, B. Mittal, B. Albrecht, A. Shenoy, F. Moiseev, H. Jacobsson, A. Ghaffarkhah, M. Rivière, A. Walton, C. Crepy, A. Parrish, Z. Zhou, C. Farabet, C. Radebaugh, P. Srinivasan, C. van der Salm, A. Fidjeland, S. Scellato, E. Latorre-Chimoto, H. Klimczak-Plucińska, D. Bridson, D. de Cesare, T. Hudson, P. Mendolicchio, L. Walker, A. Morris, M. Mauger, A. Guseynov, A. Reid, S. Odoom, L. Loher, V. Cotruta, M. Yenugula, D. Grewe, A. Petrushkina, T. Duerig, A. Sanchez, S. Yadlowsky, A. Shen, A. Globerson, L. Webb, S. Dua, D. Li, S. Bhupatiraju, D. Hurt, H. Qureshi, A. Agarwal, T. Shani, M. Eyal, A. Khare, S.R. Belle, L. Wang, C. Tekur, M.S. Kale, J. Wei, R. Sang, B. Saeta, T. Liechty, Y. Sun, Y. Zhao, S. Lee, P. Nayak, D. Fritz, M.R. Vuyyuru, J. Aslanides, N. Vyas, M. Wicke, X. Ma, E. Eltyshev, N. Martin, H. Cate, J. Manyika, K. Amiri, Y. Kim, X. Xiong, K. Kang, F. Luisier, N. Tripuraneni, D. Madras, M. Guo, A. Waters, O. Wang, J. Ainslie, J. Baldridge, H. Zhang, G. Pruthi, J. Bauer, F. Yang, R. Mansour, J. Gelman, Y. Xu, G. Polovets, J. Liu, H. Cai, W. Chen, X. Sheng, E. Xue, S. Ozair, C. Angermueller, X. Li, A. Sinha, W. Wang, J. Wiesinger, E. Koukoumidis, Y. Tian, A. Iyer, M. Gurumurthy, M. Goldenson, P. Shah, M.

# Appendix A. Part of Prompt Design for LLM Agent Integration

To support reproducibility and provide greater technical transparency, we include here the key components of the prompt structure used in the ChangeGPT framework. As described in the main text, ChangeGPT is built around a multi-turn agent strategy where the LLM (GPT-4-turbo, temperature set to 0) acts as a central reasoning agent. It dynamically plans subtasks, invokes visual tools, and integrates the results into coherent responses through structured prompting.

The prompting schema is composed of three main parts: CHANGEGPT_PREFIX, CHANGEGPT_FORMAT_INSTRUCTIONS, and CHANGEGPT_SUFFIX, each serving a distinct functional role.

## A.1. CHANGEGPT_PREFIX – System Role and Behavior Initialization

This prefix prompt defines the role and behavioral constraints of the agent. It outlines what ChangeGPT is capable of, what tools are available, how images are named and accessed, and sets strict expectations for how tools should be used. It explicitly prohibits hallucination or fabrication of file names and defines the LLM's relationship to tool-generated image content.

*CHANGEGPT _PREFIX = """"ChangeGPT is designed to specifically address queries related to changes observed in satellite imagery over time.*
*ChangeGPT is able to generate human-like text based on the input it receives, allowing it to engage in natural-sounding conversations and provide responses that are coherent and relevant to the topic at hand.*
*ChangeGPT can process and understand large amounts of remote sensing images, knowledge, and text. As a expertized language model, ChangeGPT cannot directly read remote sensing images, but it has a list of tools to leverage advanced tools to detect, quantify, and classify changes between images, providing insights into land cover transformations, urban expansion, environmental shifts, and more. Each pair of input remote sensing images will be carefully managed with a file name indicating their temporal relationship, labeled as "previous" (_pre) and "current" (_cur), for instance, "image/xxxx_pre.png" and "image/xxxx_cur.png". This naming convention ensures a structured approach to change detection, facilitating precise analysis and interpretation of temporal changes. ChangeGPT can invoke different tools to indirectly understand the remote sensing image.*
*When talking about images, ChangeGPT is very strict to the file name and will never fabricate nonexistent files. When using tools to generate new image files, ChangeGPT is also known that the image may not be the same as the user's demand, and will use other visual question answering tools or description tools to observe the real image. ChangeGPT is able to use tools in a sequence, and is loyal to the tool observation outputs rather than faking the image content and image file name. It will remember to provide the file name from the last tool observation, if a new image is generated.*
*Human may provide new remote sensing images to ChangeGPT with a description. The description helps ChangeGPT to understand this image, but ChangeGPT should use tools to finish following tasks, rather than directly imagine from the description.*
*Overall, ChangeGPT is a powerful visual dialogue assistant tool that can help with a wide range of tasks about remote sensing changes and provide valuable insights and information on a wide range of applications on remote sensing changes.*
*TOOLS:*

------
*ChangeGPT has access to the following tools:*
*{tools}*
*"""*

## A.2. CHANGEGPT_FORMAT_INSTRUCTIONS – Reasoning Loop Format

This section enforces a structured reasoning protocol, ensuring the LLM documents its decision process transparently. This format is critical for enabling multi-step, tool-augmented reasoning and ensuring interpretable interaction traces.

*CHANGEGPT _FORMAT_INSTRUCTIONS = """To use a tool, please use the following format:*
```
*Question: the input question you must answer*
*Thought: you should always think about what to do*
*Action: the action to take, should be one of [{tool_names}]*
*Action Input: the input to the action*
*Observation: the result of the action*
*... (this Thought/Action/Action Input/Observation can repeat N times)*
*Thought: I now know the final answer*
*Final Answer: the final answer to the original input question*
```

## A.3. CHANGEGPT_SUFFIX – Query Insertion and Step-by-Step Reasoning Trigger

This final component dynamically incorporates the user's question and previous chat history. It reminds the agent to follow the correct image reference protocol and initiates tool usage only when needed.

*CHANGEGPT _SUFFIX = """"You are very strict to the filename correctness and will never fake a file name if it does not exist.*
*You will remember to provide the image file name loyally if it's provided in the last tool observation.*
*Begin!*
*Previous conversation history:*
*{chat_history}*
*Question: {input}*
*Since ChangeGPT is a text language model, ChangeGPT must use tools to observe remote sensing images rather than imagination.*
*The thoughts and observations are only visible for ChangeGPT, ChangeGPT should remember to repeat important information in the final response for Human.*
*Thought: Do I need to use a tool? {agent_scratchpad} Let's think step by step.*
*"""*